\documentclass[journal]{IEEEtran}
\usepackage{amsmath,amsfonts,amssymb}
\usepackage{algorithmic}
\usepackage{algorithm}
\usepackage{makecell}
\usepackage{array}
\usepackage[caption=false,font=normalsize,labelfont=sf,textfont=sf]{subfig}
\usepackage{textcomp}
\usepackage{stfloats}
\usepackage{url}
\usepackage{pifont}
\newcommand{\cmark}{\ding{51}} 
\newcommand{\xmark}{\ding{55}} 
\usepackage{graphicx} 
\usepackage{amssymb}
\usepackage{verbatim}
\usepackage{graphicx}
\usepackage{cite}
\usepackage{xspace}
\usepackage{xcolor}
\usepackage[table]{xcolor}
\usepackage{color}
\usepackage{bbding}
\usepackage{lineno}
\usepackage{booktabs} 
\usepackage{multirow}
\usepackage{arydshln}
\usepackage{enumitem}
\usepackage{hyperref}

\definecolor{MyDeepGreen}{RGB}{0,150,0} 

\definecolor{MyDeepBlue}{RGB}{50,150,240} 

\begin{document}

\title{Salient Object Detection in Complex Weather Conditions via Noise Indicators}

\author{Quan Chen, Xiaokai Yang, Tingyu Wang, Rongfeng Lu, Xichun Sheng, Yaoqi Sun,\\ Chenggang Yan
\thanks{Quan Chen is with the School of Automation, Hangzhou Dianzi University, Hangzhou 310018, China, and also with the College of Artificial Intelligence, Jiaxing University, Jiaxing 314001, China~(e-mail:chenquan@alu.hdu.edu.cn).}
\thanks{Xiaokai Yang and Tingyu Wang are with the School of Communication Engineering, Hangzhou Dianzi University, Hangzhou 310018, China~(e-mail:242080082@hdu.edu.cn, tingyu.wang@hdu.edu.cn).}
\thanks{Rongfeng Lu and Chenggang Yan are with the School of Automation, Hangzhou Dianzi University, Hangzhou 310018, China~(e-mail:rongfeng-lu@hdu.edu.cn, cgyan@hdu.edu.cn).}
\thanks{Xichun Sheng is with the School of Faculty of Applied Science, Macao Polytechnic University, Macao, China~(email:p2314922@mpu.edu.mo)}
\thanks{Yaoqi Sun is with the School of Mathematics and Computer Science, Lishui University, Lishui 323000, and also with the Lishui Institute of Hangzhou Dianzi University, Lishui 323050, China~(email:sunyq2233@163.com)}
}

\markboth{Journal of \LaTeX\ Class Files,~Vol.~14, No.~8, August~2021}%
{Shell \MakeLowercase{\textit{et al.}}: A Sample Article Using IEEEtran.cls for IEEE Journals}


\maketitle

\begin{abstract}
Salient object detection~(SOD), a foundational task in computer vision, has advanced from single-modal to multi-modal paradigms to enhance generalization. However, most existing SOD methods assume low-noise visual conditions, overlooking the degradation of segmentation accuracy caused by weather-induced noise in real-world scenarios. In this paper, we propose a SOD framework tailored for diverse weather conditions, encompassing a specific encoder and a replaceable decoder. To enable handling of varying weather noises, we introduce a one-hot vector as a noise indicator to represent different weather types and design a Noise Indicator Fusion Module~(NIFM). The NIFM takes both semantic features and the noise indicator as dual inputs and is inserted between consecutive stages of the encoder to embed weather-aware priors via adaptive feature modulation. Critically, the proposed specific encoder retains compatibility with mainstream SOD decoders. Extensive experiments are conducted on the WXSOD dataset under varying training data scales (100\%, 50\%, 30\% of the full training set), three encoder and seven decoder configurations. Results show that the proposed SOD framework~(particularly the NIFM-enhanced specific encoder) improves segmentation accuracy under complex weather conditions compared to a vanilla encoder.
\end{abstract}


\section{Introduction}
\IEEEPARstart{S}{alient} object detection~(SOD), a fundamental task in computer vision, aims to mimic the attention mechanism of the human visual system to accurately locate and highlight the most visually attractive regions in complex scenes. 
As a core technology supporting downstream applications, SOD has long remained a research focus. With the advancement of studies, this field has achieved a leap from single-modality RGB SOD~\cite{zhuge2019deep,wang2023pixels,jin2024underwater,song2024universal,zhang2020multi,he2025samba} to multi-modality paradigms, including RGB-Thermal~\cite{wang2021cgfnet,liu2021swinnet,zhou2023wavenet}, RGB-Depth~\cite{chen2022modality,zhang2023multi,wang2024learning}, even triple-modal RGB-Depth-Thermal SOD~\cite{wan2023mffnet,luo2024dynamic,bao2025ifenet}. These advancements enhance the generalization ability and scene adaptability of SOD models.

\begin{figure}[!t]
\centering
\includegraphics[width=1.0\linewidth]{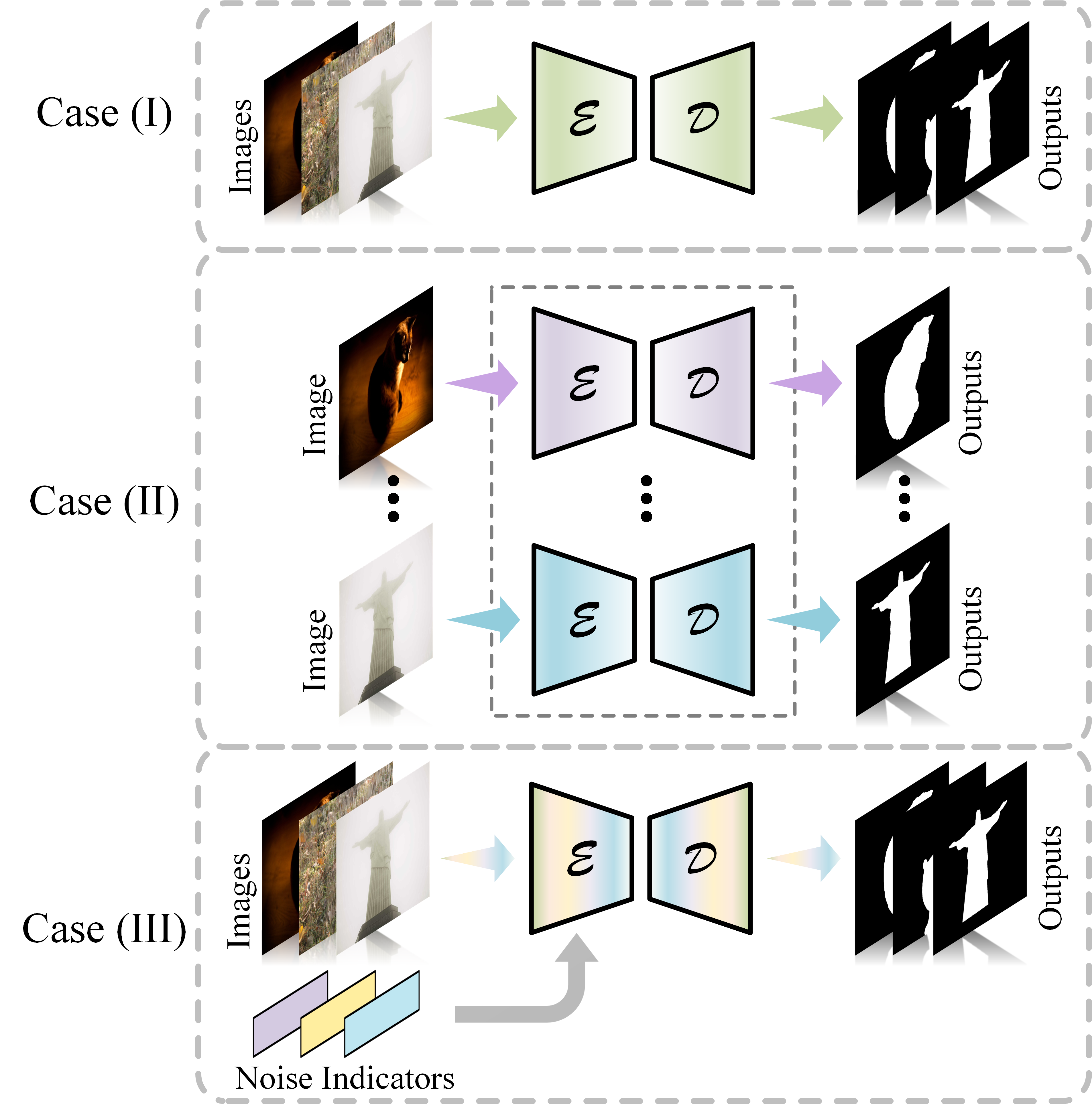}
\caption{Three paradigms for salient object detection under complex weather conditions.
Case (I): A unified network is used for both training and testing, without accounting for the variations in weather-induced noise. Case (II): A distinct model is trained independently for images with different weather noise, incurring significant computational redundancy. Case (III): Our proposed framework, which introduces a noise indicator to highlight the differences between various weather noises, thereby enhancing the network’s detection robustness across diverse weather conditions.
}
\label{basicflow}
\end{figure}  

Despite these notable advances, most existing SOD methods assume that input images are under ideal visual conditions (\textit{i.e.}, noise-free or low-noise environments), ignoring the impact of weather-induced noise on segmentation accuracy in real-world scenarios. With the rapid expansion of computer vision into practical applications, research focus has shifted toward algorithm design for low-quality scenarios, a trend that has further prompted the development of unified models to enhance robustness under diverse weather conditions.
For instance, in tasks such as image restoration~\cite{li2022all,potlapalli2023promptir,jiang2024autodir} and retrieval~\cite{wang2024multiple,feng2024multi,wen2025weatherprompt}, several studies have built unified frameworks to mitigate weather-related distortions. 
These efforts motivate us to develop a SOD framework capable of handing complex weather-corrupted inputs while maintaining compatibility with prevailing SOD methods.

We first analyzes two typical strategies to process diverse weather-corrupted images, alongside the proposed general strategy, as illustrated in Fig.~\ref{basicflow}.
Case~(I): Images with various weather noises are utilized for training and testing without differentiation. This strategy entirely neglects the inherent variations between different weather noises types, resulting to constrained model adaptability in complex conditions.
Case~(II): A specialized model is trained for each weather type. Although tailored to specific noise patterns, it requires storing multiple sets of model parameters, inevitably increasing computational costs.
Case~(III): The proposed general strategy guides the encoder to learn discriminative feature representations related to noise via noise indicators, while ensuring compatibility with a wide range of existing decoders. 
Compared with the first two cases, our strategy combines the advantages of both: it employs a single encoder for architectural simplicity, yet enhances discriminative capability across weather types through explicit noise-aware conditioning.

To implement this general strategy, we decompose the SOD framework into core components: a specific encoder and a replaceable decoder, with our design efforts focused on the specific encoder. Specifically, we introduce a one-hot vector to represent distinct weather noise types, and propose a Noise Indicator Fusion Module~(NIFM) that accepts both semantic features and the one-hot vector as dual inputs. 
NIFM is inserted between consecutive stages of the backbone network to progressively embed weather priors, thereby guiding the encoder to learn unified and robust multi-scale semantic features through adaptive feature modulation.
Crucially, the specific encoder retains compatibility with mainstream SOD decoders, enabling direct integration into existing SOD pipelines without requiring structural adjustments.
Experimental results on the WXSOD dataset show that the proposed SOD framework improve segmentation accuracy under complex weather conditions across varying training data scales, backbone and decoder architectures, providing a feasible pathway toward diverse-weather SOD.
Our primary contributions are summarized as follows:
\begin{itemize}
\item We propose a SOD framework tailored to diverse weather conditions, comprising a specific encoder and a replaceable decoder. 
\item We design a NIFM taking semantic features and a noise indicator as dual inputs and guiding the encoder to learn robust feature representations
\item Experiments results on the WXSOD dataset, as well as under different backbone and decoder configurations, have verified the effectiveness of our method.
\end{itemize}

The rest of the paper is organized as follows: Section~\ref{Related Works} presents related work, focusing on single-modality and multi-modality SOD. In Section~\ref{Methods}, we detail our proposed method, including the method overview and each component. Section~\ref{Experiments and Results} presents our experimental results, and Section~\ref{Conclusions} provides concluding remarks.

\section{Related Work}\label{Related Works}

\subsection{RGB SOD} 
Salient Object Detection has garnered significant research interest in recent years, leading to the proposal of numerous SOD methods. Traditional methods~\cite{jiang2013salient,kim2015salient} primarily relied on handcrafted low-level visual features, such as color contrast, edge information, and texture consistency. Although these models perform adequately in simple backgrounds, they are susceptible to interference and exhibit poor robustness when confronted with complex scenes.

In contrast, convolutional neural networks~(CNN) have achieved remarkable success in a wide range of computer vision tasks~\cite{yu2025state,10587023,chen2024scale,feng2024u2udata,feng2025evoagent,feng2025embodied,yu2025crisp,yu2025visual,shi2025multi,he2025temporal,liu2025yolo,guo2025selective}. Researchers have proposed numerous CNN-based SOD models, which are underpinned by diverse theoretical paradigms such as visual attention mechanisms~\cite{chen2018reverse}, the interaction of multi-level features~\cite{pang2020multi} and the utilization of edge information~\cite{wang2022multiscale}. These innovations have considerably advanced SOD performance. However, current SOD datasets~\cite{yang2013saliency,shi2015hierarchical,wang2017learning,deng2023recurrent,zhao2024local} neglect noise interference, resulting in learning-based methods being sensitive to noise. To address this gap, Wan~\textit{et al.}~\cite{wan2024adnet} introduced salt-and-pepper noise with varying intensities to guide models in learning noise-robust feature representations. To mitigate the effects of image dark degradation and low contrast, Yu~\textit{et al.}~\cite{yu2024degradation} combined SOD networks with low-light enhancement techniques to form a novel learning framework, while constructing the first dedicated SOD dataset for low-light scenarios. These datasets suffer from limited noise diversity and fail to real-world complexity. Consequently, Yuan~\textit{et al.}~\cite{yuan2024bi} made a preliminary attempt to incorporate weather noise into ORSI SOD, adding fog and rain noise to enhance the generalization of the model, but did not construct a benchmark. Chen~\textit{et al.}~\cite{chen2025wxsod} developed the WXSOD, the first large-scale SOD dataset incorporating both diverse synthetic and real-world noises under adverse weather conditions. Differ from specific-structured models, our objective is to develop a simple yet effective framework capable of enhancing the saliency detection accuracy of various methods under extreme weather conditions.

\subsection{Multi-modal SOD}
As complementary information to RGB images, depth and thermal modalities have been widely adopted in SOD tasks. 
These complementary properties have spurred rapid advancements in multi-modality SOD, leading to specialized sub-tasks, including RGB-D SOD, RGB-T SOD, and RGB-DT SOD.

In the RGB-D SOD domain, research focuses on exploring correlations and disparities between RGB and depth features~\cite{chen2020dpanet,wen2021dynamic,yao2023depth,li2023delving,cong2023point,zhou2025location}.For instance, Piao~\textit{et al.}~\cite{piao2020a2dele} introduced a depth distiller to transfer depth knowledge from the depth stream to the RGB stream, eliminating redundant depth information. 
To mitigate this limitation, recent studies emphasize optimized feature fusion strategies, which can be divided into two categories: i) Pixel-level fusion: This approach concatenates RGB and depth images into four-channel inputs processed by a single-stream network. 
ii) Feature-level fusion: As the predominant paradigm in current RGB-D SOD research, this strategy extracts uni-modal features via dual-branch networks before integration. For instance, Liu~\textit{et al.}~\cite{liu2021learning} enhanced cross-modal interactions through mutual attention combined with contrast learning and adaptive selective attention.
In the parallel field of RGB-T SOD, Tu~\textit{et al.}~\cite{tu2022weakly} planned a large-scale RGB-T dataset with degraded samples and developed an end-to-end network for adaptive salient cue selection. 
From the perspective of network structure optimization, it is worth noting that the technical advancements developed for RGB-D SOD models and RGB-T SOD models exhibit significant interconnections~\cite{cong2022does,zhou2023wavenet,zhou2023position,zhou2024frequency,tang2024divide,wang2024alignment,wang2025alignment}.
To transcend dual-modality constraints, RGB-DT SOD integrates three modalities to enhance saliency detection accuracy in complex scenarios~\cite{song2022novel,wan2023mffnet,luo2024dynamic,bao2024quality,bao2025ifenet}. 
Toward enhanced deployment flexibility, recent advances leverage~\cite{huang2024salient,huang2024modality,wang2025unified} leverage modality-type indicators or prompt learning to construct SOD frameworks compatible with different modality inputs. 
For instance, Huang~\textit{et al.}~\cite{huang2024salient} incorporated one-hot encoding of image modality types into the model architecture, thereby enhancing the accuracy and generalization of multi-modal SOD. 
Inspired by these developments, we posit that explicitly distinguishing noise types of RGB images captured under diverse weather conditions is promising to enhance SOD robustness against varying noise interference.
Based on this insight, this paper introduces a novel noise-prompted SOD framework designed for complex weather conditions.

\begin{figure*}[!t]
\centering
\includegraphics[width=1.0\linewidth]{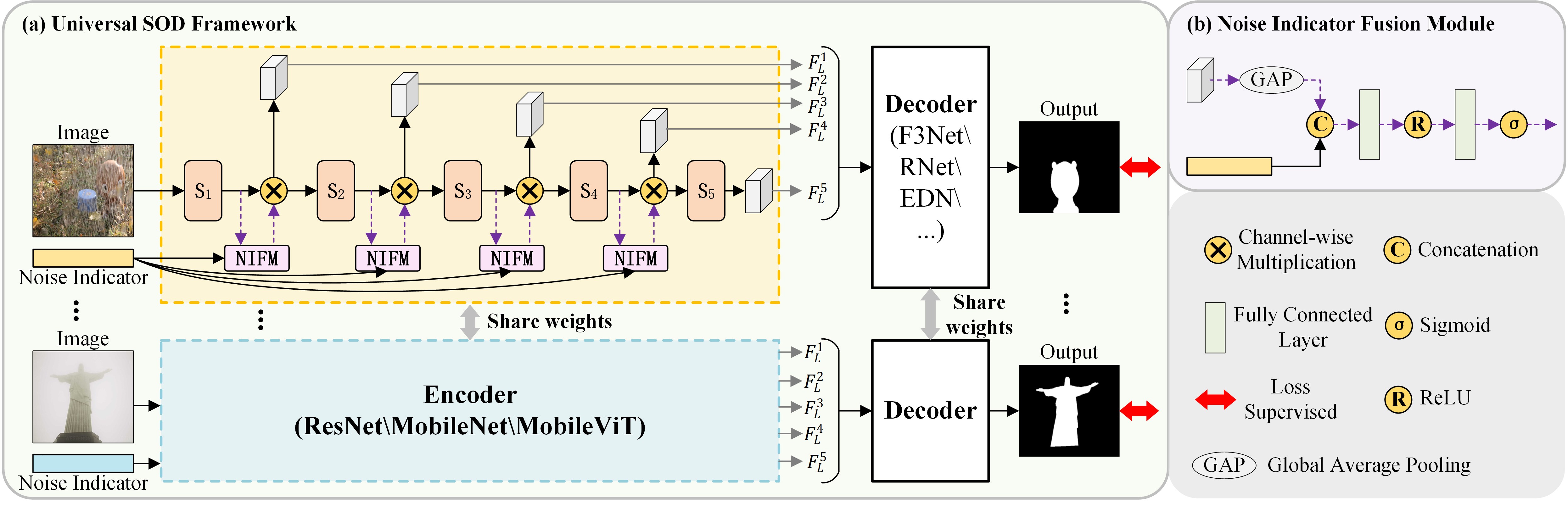}
\caption{The diagram of the proposed framework. The input image and noise indicator are first fed into the specific encoder to extract multi-scale semantic features. Then, the replaceable decoder is adopted to map semantic features into salient images. Note that we employs 7 various decoders to verify the feasibility of the proposed framework.}
\label{model}
\end{figure*}  


\subsection{Object detection in Adverse Weather Conditions}
Recent studies on object detection have focused on improving detection accuracy under adverse weather conditions. 
These methods can be classified into three categories according to their implementation workflows.
The first category~\cite{gupta2024robust} directly trains models on images captured under real or simulated adverse weather. 
Although this strategy improves model performance in complex conditions, it requires constructing datasets covering diverse adverse weather scenarios.
The second category~\cite{zhang2023adaptive} employs a two-stage workflow that first applies image restoration methods~\cite{wu2021contrastive} to denoise images and then feeds the restored images into object detection models for detection. 
While these methods improve image clarity, their performance generally lags behind the first category, largely because image restoration process may alter-intrinsic the original features of images. 
The third category~\cite{li2023detection} integrates weather-aware image enhancement techniques into the detection framework or utilizes domain adaptation to minimize the domain shift between clear-weather training data and adverse-weather test data. 
Although these approaches often surpass the second category in performance, studies~\cite{li2024ia,wang2022r} indicate that their detection results are less consistent than those achieved by models trained directly on adverse weather images.
Inspired by the aforementioned works, we propose a novel framework to achieve salient object detection in complex weather scenarios.

\section{Methodology}\label{Methods}
In this section, we detail the proposed network. In Section~\ref{Overview}, we give an overview of the basic network architecture. In Section~\ref{Noise-indicator Fusion Module}, we present the Noise Indicator Fusion Module in detail. After that, in Section~\ref{Decoders}, we have provided details of various decoders. Finally, in Section~\ref{Discussion}, we further discuss the basic network framework.

\subsection{Overview}\label{Overview}
The overview of our framework is shown in Fig.~\ref{model}. Given an input image ${I}_{in}\in\mathbb{R}^{H \times W\times C}$ with a known noise type, we first construct a noise indicator ${T}\in\mathbb{R}^{M}$ that uniquely encodes the noise type. This one-hot representation corresponds to $M$ predefined weather-induced noise categories. Both ${I}_{in}$ and ${T}$ are then fed into a ResNet-50 backbone to extract multi-scale semantic features $\{F_{i}\}_{i=1}^{5}$. To enable interaction between noise indicator and semantic features, a Noise Indicator Fusion Module~(NIFM) is inserted between every two consecutive stages of ResNet-50. Then, a replaceable decoder is employed to map the multi-scale semantic features into a saliency map. It is worth noting that we focus on designing a general framework rather than a specific network structure. Therefore, we implement our framework with 7 various decoders (including F3Net~\cite{wei2020f3net}, R-Net~\cite{wang2025r}, MINET~\cite{pang2020multi}, AESINet~\cite{zeng2023adaptive}, BAFSNet~\cite{gu2023orsi}, DC-Net~\cite{zhu2025dc} and EDN~\cite{wu2022edn}) for comprehensive validation. 
Decoder implementations follow original configurations as described in the cited publications.

\subsection{Noise-indicator Fusion Module}\label{Noise-indicator Fusion Module}
Studies addressing weather noise variations in other visual tasks often attempt to explicitly incorporate weather-type encoding vectors~\cite{serrano2024adaptive,rajagopalan2025gendeg} or insert learnable vectors~\cite{gao2024prompt,wu2025learning}, thereby guiding networks to distinguish differences in image noise. 
Meanwhile, emerging studies~\cite{huang2024salient,huang2024modality,wang2025unified} in SOD focus on designing multi-modal unified frameworks, employing similar approaches to empower the model to discern and adapt to different input data modalities.
Inspired by these efforts, we design a Noise Indicator Fusion Module including a noise indicator.
Assuming there are $M$ types of weather-noise, the noise indicator $T\in\mathbb{R}^{M}$ is a one-hot encoded vector of length $M$, where the element corresponding to the noise type of the input image is set to 1, and other elements are set to 0.
For instance, the noise indicator for rain images is $[0,1,0,0,0,0,0,0,0]$, whereas that corresponding to snow images is $[0,0,1,0,0,0,0,0,0]$.
Next, we elaborate on the usage of the predefined one-hot noise indicators.

Taking a RGB image $I_{in}$ as input, ResNet-50 maps it into high-dimensional feature representation:
\begin{align}\label{eq1}
\{\hat{F}_{1},\hat{F}_{2},\hat{F}_{3},\hat{F}_{4},\hat{F}_{5}\} = \mathcal{T}_{resnet}(I_{in})
\end{align}
where $\mathcal{T}_{resnet}(\cdot)$ denotes the feature extraction operation, and $\hat{F}_{i}$ denotes the semantic feature of the $i$-th stage~($S_{i}$) extracted by the pure ResNet-50.

To integrate this noise indicator into the backbone network, we design a dedicated module termed the noise indicator fusion module~(NIFM).
As illustrated in Fig.~\ref{model}, NIFMs are inserted between every two adjacent stages of the backbone network. 
Taking the first NIFM~(positioned after the backbone’s first stage, $S_1$) as an example, the inputs consist of two components: the semantic feature  
$F_{1}$ extracted from the input image $I_{in}$ after $S_{1}$ of the backbone network, and the pre-constructed noise indicator $T$. 
The semantic feature $F_{1}$ is compressed into a one-dimensional vector via a global average pooling layer~(GAP), which is then concatenated with the noise indicator to form a combined feature representation:
\begin{equation}
   F_{1}^{T} = Concat(GAP(\hat{F}_{1}),T)
\end{equation}
where $Concat(\cdot,\cdot)$ denotes the channel-wise concatenation operation. Next, the concatenated feature $F_{1}^{T}$ is fed into two fully connected layers~($fc$) to learn a noise indicator weight $W_{1}$. This weight vector encodes the degree to which the backbone’s features should be adjusted to counteract the specific noise type indicated by $T$.
The weight calculation is formulated as:
\begin{equation}
    W_{1} = \sigma(fc_{2}(relu(fc_{1}(F_{1}^{T}))))
\end{equation}
where $\sigma(\cdot)$ denotes the sigmoid operation. 
Finally, the learned weight $W_{1}$ is applied to modulate the backbone network, thereby incorporating noise-aware conditioning and producing the enhanced output feature $F_{1}$ of the first stage:
\begin{equation}
    F_{1}=W_{1}\otimes\hat{F}_{1}
\end{equation}
where $\otimes$ denotes the channel-wise multiplication. 
It can be seen that $W_{1}$ implements the switching of noise types for the features $\hat{F}_{1}$.
The same operational logic extends to all subsequent NIFMs. For the $n$-th NIFM (where $n = 2, 3, 4$), the inputs are the semantic feature $\hat{F}_n$ (extracted from the backbone’s $n$-th stage, $S_n$) and the noise indicator $T$.
In this manner, both the noise-specific features and their associated indicator weights are progressively derived for the remaining stages. 
From Eq.~(\ref{eq1}), we can formulate the multi-scale semantic feature extraction of the NIFM-integrated backbone network as:
\begin{equation}
\{F_{1},F_{2},F_{3},F_{4},F_{5}\}=\mathcal{T}_{resnet}^{NIFM}(I_{in})
\end{equation}
where $\mathcal{T}_{resnet}^{NIFM}(\cdot)$ denotes the NIFM-augmented ResNet-50, \textit{i.e.}, the specific encoder.

\subsection{Decoders}\label{Decoders}
Mainstream backbone-based SOD methods can generally be decomposed into two core components: an encoder and a decoder. 
The encoder is engineered to extract multi-scale semantic features, while decoders, distinguished by diverse architectural designs (\textit{e.g.}, pyramid feature fusion, cross-scale attention, or residual refinement modules), are tasked with progressively mapping these hierarchical semantic features into dense, pixel-wise saliency maps.
To validate the effectiveness and generalization of the proposed specific encoder, we conduct experiments by pairing this encoder with decoders from seven representative SOD models that cover different architectural paradigms. 
These models include F3Net~\cite{wei2020f3net}, R-Net~\cite{wang2025r}, MINET~\cite{pang2020multi}, AESINet~\cite{zeng2023adaptive}, BAFSNet~\cite{gu2023orsi}, DC-Net~\cite{zhu2025dc} and EDN~\cite{wu2022edn}. 
To ensure a fair comparison, all models adopt the same loss function and retain original multi-scale supervision mechanisms. Taking the single-scale supervision as an example, the loss function can be formulated as follows:
\begin{equation}
\mathcal{L}_{total}=\mathcal{L}_{bce}+\mathcal{L}_{ssim}+\mathcal{L}_{iou}
\end{equation}
where $\mathcal{L}_{bce}$, $\mathcal{L}_{ssim}$ and $\mathcal{L}_{iou}$ denote BCE loss~\cite{de2005tutorial}, SSIM loss~\cite{wang2003multiscale} and IoU loss~\cite{mattyus2017deeproadmapper}, respectively.
The calculation process of each loss function is as follows:
\begin{equation}
\left\{\begin{array}{l}
\mathcal{L}_{bce}=-I_{gt}\cdot\text{log}(I_{out})+(I_{gt}-1)\cdot\text{log}(1-I_{out})\\
\mathcal{L}_{ssim}=1-\frac{(2\mu_{I_{gt}}\mu_{I_{out}}+C_{1})(2\sigma_{I_{gt}I_{out}}+C_{2})}{(\mu_{I_{gt}}^{2}+\mu_{I_{out}}^{2}+C_{1})(\sigma_{I_{gt}}^{2}+\sigma_{I_{out}}^{2}+C_{2})} \\
\mathcal{L}_{iou}=1-\frac{I_{gt}\cdot I_{out}}{I_{gt}+I_{out}-I_{gt}\cdot I_{out}}
\end{array}\right.
\end{equation}
where $I_{out}$ and $I_{gt}$ denote the predicted saliency map and ground-truth. $\mu_{I_{out}}$, $\mu_{I_{gt}}$ and $\sigma_{I_{out}}$, $\sigma_{I_{gt}}$ are the mean and standard deviations of $I_{out}$ and $I_{gt}$ respectively, $\sigma_{I_{gt}I_{out}}$ is their covariance, $C_1=0.012$ and $C_2=0.032$ are used to avoid dividing by 0.

\subsection{Discussion}\label{Discussion}
With the rapid advancement of computer vision, researchers have devoted growing attention to the algorithmic application for low-quality scenarios.
These efforts have motivated us to develop a frameworks for achieving robust SOD under diverse weather conditions.
In parallel, recent progress in the SOD field has explored the settings of various modal inputs to improve the generalization capability of models. 
To realize this, existing methods incorporate modality indicators or learnable prompts to guide a single backbone network in adapting to input images of multiple modalities, thereby eliminating the need for separate network architectures for different modalities. 
Such technical designs provide valuable insights for our research. 
However, a fundamental distinction must be clarified: images captured under complex weather conditions~(\textit{e.g.}, rain, fog or snow) inherently belong to the same visual modality. 
This implies that the conventional SOD model can technically process weather-corrupted images without additional modifications, albeit with suboptimal performance. 
Therefore, the focus of our work is the design of a encoder. 
Concretely, we employ a simple yet effective NIFM to enhance the backbone network's capability of extracting features from images corrupted by different types of weather-induced noise, while ensuring compatibility with existing decoders.
The experimental section will verify that introducing noise indicators can optimize the spatial distribution of images with different types of noise, promoting the mutual proximity of image features with the same noise category.

\begin{table*}[!t]
\renewcommand{\arraystretch}{1.25}
\centering
\caption{Statistics of the WXSOD Dataset.}
\label{datasets}
\resizebox{1.0\linewidth}{!}{
\begin{tabular}{ccccccccccccccccc}
\hline
\multicolumn{17}{c}{WXSOD}    \\ \hline
\multicolumn{9}{c}{100\% Training set}   &  & \multirow{2}{*}{50\% Training set} &  & \multirow{2}{*}{30\% Training set} &  & \multirow{2}{*}{Synthesized test set} &  & \multirow{2}{*}{Real test set} \\ \cline{1-9}
Clean & Rain & Snow & Fog  & Light & Dark & Rain\&Snow & Rain\&Fog & Snow\&Fog &  &         &  &       &  &               &  &         \\ \cline{1-9} \cline{11-11} \cline{13-13} \cline{15-15} \cline{17-17} 
631   & 1524 & 1547 & 1534 & 1531  & 1535 & 1494       & 1562      & 1533      &  & 6445       &  & 3867          &  & 1500      &  & 554      \\ \hline
\end{tabular}
}
\end{table*}  

\begin{table*}[!t]
\renewcommand{\arraystretch}{1.10}
\centering
\caption{Model performance comparisons using 100\% training set. The upper section presents results on the Synthesized test set, while the lower section reports results on the real test set.
Four models~(A3Net~\cite{cui2023autocorrelation}, HDNet~\cite{lu2024low}, ICONet~\cite{zhuge2022salient}, TCGNet~\cite{liu2023tcgnet}) with purple background color adopt default structure.}
\label{results_100}
\resizebox{\linewidth}{!}{
\begin{tabular}{c|c|c|cc|cc|cc|cc|cc|cc|cc|cc|ccc}
\hline
\multirow{2}{*}{Test set} &
\multirow{2}{*}{Methods} &
\multirow{2}{*}{NIFM} &
\multicolumn{2}{c|}{${MAE} \downarrow$} &
\multicolumn{2}{c|}{${S} \uparrow$} &
\multicolumn{2}{c|}{${F}_{\beta}^{adp} \uparrow$} &
\multicolumn{2}{c|}{${F}_{\beta}^{mean} \uparrow$} &
\multicolumn{2}{c|}{${F}_{\beta}^{max} \uparrow$} &
\multicolumn{2}{c|}{${E}_{\xi}^{adp} \uparrow$} &
\multicolumn{2}{c|}{${E}_{\xi}^{mean} \uparrow$} &
\multicolumn{2}{c|}{${E}_{\xi}^{max} \uparrow$} & \multirow{2}{*}{MACs(G)}& \multirow{2}{*}{Params(M)}& \multirow{2}{*}{FPS} \\
& &  & val & $\Delta$ & val & $\Delta$ & val & $\Delta$ & val & $\Delta$ & val & $\Delta$ & val & $\Delta$ & val & $\Delta$ & val & $\Delta$ \\ \hline
\multirow{19}{*}{Sys} &  \cellcolor[HTML]{E7E9FF}A3Net   & \cellcolor[HTML]{E7E9FF}-- & \cellcolor[HTML]{E7E9FF}0.0327 & \cellcolor[HTML]{E7E9FF}--      & \cellcolor[HTML]{E7E9FF}0.8784 & \cellcolor[HTML]{E7E9FF}--      & \cellcolor[HTML]{E7E9FF}0.8158 & \cellcolor[HTML]{E7E9FF}--      & \cellcolor[HTML]{E7E9FF}0.8286 & \cellcolor[HTML]{E7E9FF}--      & \cellcolor[HTML]{E7E9FF}0.8472 & \cellcolor[HTML]{E7E9FF}--      & \cellcolor[HTML]{E7E9FF}0.9176 & \cellcolor[HTML]{E7E9FF}--      & \cellcolor[HTML]{E7E9FF}0.9131 & \cellcolor[HTML]{E7E9FF}--      & \cellcolor[HTML]{E7E9FF}0.9253 & \cellcolor[HTML]{E7E9FF}-- &  \cellcolor[HTML]{E7E9FF}22.70 & \cellcolor[HTML]{E7E9FF}27.94 & \cellcolor[HTML]{E7E9FF}228.08 \\
\hline
\end{tabular}
}
\end{table*} 

\section{Experiment}\label{Experiments and Results}

\subsection{Datasets and Evaluation Metrics}\label{Datasets and Evaluation Metrics}
\subsubsection{Datasets}
Experiments are conducted on the WXSOD dataset~\cite{chen2025wxsod}, a dedicated SOD benchmark for adverse weather conditions. As detailed in Table~\ref{datasets}, the original WXSOD comprises 12,891 images for training, 1,500 synthesized and 554 real images for testing. Notably, the 12,891 training samples can be further categorized into 9 subsets based on weather conditions, including Clean~(631 images), Rain~(1524 images), Snow~(1547 images), Fog~(1534 images), Light~(1531 images), Dark~(1535 images), Rain\&Snow~(1494 images), Rain\&Fog~(1562 images), and Snow\&Fog~(1533 images). It is emphasized that the scenes within each subset are unique. By using subsets for training, we can evaluate the model detection performance under specific weather noise.
To further verify the effectiveness of our framework across datasets of varying scales, we generate two additional training subsets by randomly filtering the original training set at two different ratios: 50\% and 30\%. These filtered subsets are denoted as the “50\% Training set” (containing 6,445 images) and the “30\% Training set” (containing 3,867 images), respectively. This experimental setup allows us to analyze how the performance of the framework changes as the amount of training data decreases, thereby evaluating the effectiveness of the framework.

\subsubsection{Evaluation Metrics}
Adhering to WFANet~\cite{chen2025wxsod}, we employ ten commonly used evaluation metrics, which are listed as follows: mean absolute error~($MAE$)~\cite{perazzi2012saliency}, S-measure~($S$)~\cite{fan2017structure}, F-measure~($F_{\beta}^{adp}, F_{\beta}^{mean}, F_{\beta}^{max}$)~\cite{achanta2009frequency}, E-measure~($E_{\xi}^{adp}, E_{\xi}^{mean}, E_{\xi}^{max}$)~\cite{fan2018enhanced}, F-measure curve and Precision-Recall~(PRcurve)~\cite{achanta2009frequency}.
These metrics evaluate the model's performance from multiple aspects, such as pixel-level error, structural consistency, and overall perceptual quality, thereby ensuring the comprehensiveness and reliability of the evaluation results.

To quantify additional computational costs incurred by the combinations of various decoders and the basic network, we report the number of learnable parameters~(Params), multiply–accumulate operations~(MACs) and frames per second~(FPS).
Note that the MACs and FPS are measured on a $384\times384$ image, using an NVIDIA P100 GPU.

\subsection{Implementation Details}\label{Model Settings and Training Details}
The basic framework employs a ResNet-50 as the encoder, with the NIFM embedded into adjacent stages of the encoder. In contrast, the decoder leverages existing network architectures.
During the training phase, the Adam optimizer~($\beta_{1}=0.9$ and $\beta_{2}=0.999$) is utilized to update network parameters with an initial learning rate of 0.001. To ensure stable convergence, the StepLR learning rate scheduling strategy is employed, where the learning rate is decayed by a factor of $\gamma=0.2$ every 40 epochs. All input images are uniformly resized to $384\times384$ resolution and processed in mini-batches of 4 over 52 epochs. The framework adopts a unified loss function when applying different decoders, but retains multi-scale loss supervision that reported in the corresponding original literature.

\begin{figure*}[!t]
\centering
\includegraphics[width=1.0\linewidth]{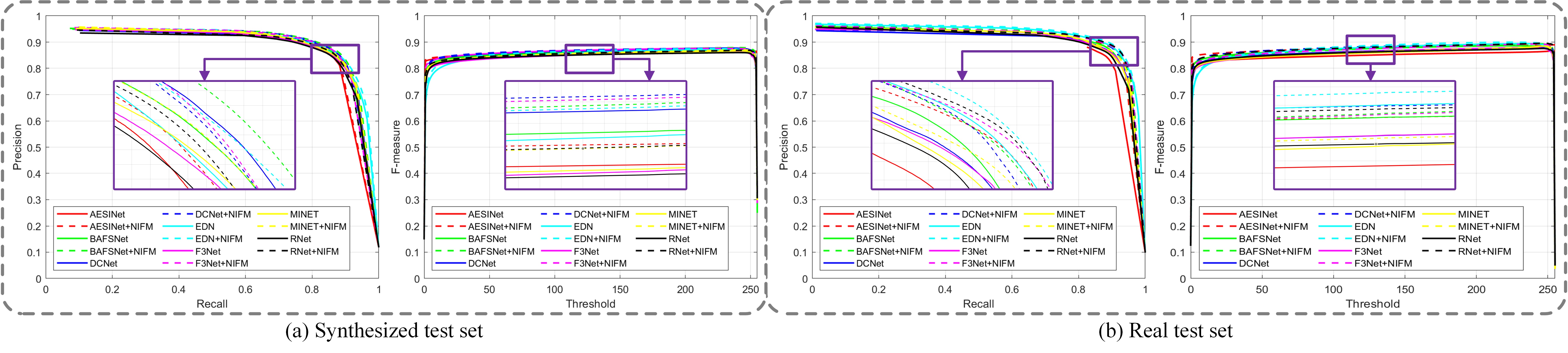}
\caption{Quantitative evaluation of different saliency models: (a) presents PR curves and F-measure curves on the synthesized test set, while (b) presents PR
curves and F-measure curves on the real test set.}
\label{PRcurve}
\end{figure*}  

\subsection{Comparison with SOTA Models}\label{Comparison with SOTA Models}

\subsubsection{Quantitative Comparison}
To comprehensively validate the effectiveness of the proposed framework, particularly the specific encoder integrated with the NIFM, quantitative experiments are conducted from two perspectives: 1) the effectiveness of the proposed framework when paired with different decoder configurations; 2) the effectiveness of the proposed framework under varying dataset scales.

\begin{table*}[!t]
\renewcommand{\arraystretch}{1.10}
\centering
\caption{Model performance comparisons using 50\% and 30\% training set, respectively. The upper section presents results on the 50\% test set, while the lower section reports results on the 30\% test set.}
\label{results5030}
\resizebox{1.0\linewidth}{!}{
\begin{tabular}{c|c|c|c|cc|cc|cc|cc|cc|cc|cc|cc}
\hline
\multirow{2}{*}{Ratio} & \multirow{2}{*}{Test set}&
\multirow{2}{*}{Decoders} &
\multirow{2}{*}{NIFM} &
\multicolumn{2}{c|}{${MAE} \downarrow$} &
\multicolumn{2}{c|}{${S} \uparrow$} &
\multicolumn{2}{c|}{${F}_{\beta}^{adp} \uparrow$} &
\multicolumn{2}{c|}{${F}_{\beta}^{mean} \uparrow$} &
\multicolumn{2}{c|}{${F}_{\beta}^{max} \uparrow$} &
\multicolumn{2}{c|}{${E}_{\xi}^{adp} \uparrow$} &
\multicolumn{2}{c|}{${E}_{\xi}^{mean} \uparrow$} &
\multicolumn{2}{c}{${E}_{\xi}^{max} \uparrow$} \\
& & &  & val & $\Delta$ & val & $\Delta$ & val & $\Delta$ & val & $\Delta$ & val & $\Delta$ & val & $\Delta$ & val & $\Delta$ & val & $\Delta$ \\ \hline
\multirow{32}{*}{{50\%}} & \multirow{16}{*}{{Sys}}     & \multirow{2}{*}{F3Net}    &   \xmark                    &    0.0391       & --         &  0.8640         &  --         &  0.7946         &   --        &  0.8060         &  --         & 0.8240          & --          & 0.9033          & --          &  0.8991         & --          &  0.9102         &  --         \\
&        &           &   \cellcolor[HTML]{E2FFFE} \cmark           &     \cellcolor[HTML]{E2FFFE}0.0343    &   \cellcolor[HTML]{E2FFFE}12.28\%        &  \cellcolor[HTML]{E2FFFE}0.8737         &  \cellcolor[HTML]{E2FFFE}1.13\%         &   \cellcolor[HTML]{E2FFFE}0.8133        &    \cellcolor[HTML]{E2FFFE}2.35\%       &    \cellcolor[HTML]{E2FFFE}0.8219       &    \cellcolor[HTML]{E2FFFE}1.97\%       &   \cellcolor[HTML]{E2FFFE}0.8399        &   \cellcolor[HTML]{E2FFFE}1.93\%        &   \cellcolor[HTML]{E2FFFE}0.9150        &   \cellcolor[HTML]{E2FFFE}1.30\%        &    \cellcolor[HTML]{E2FFFE}0.9098       &     \cellcolor[HTML]{E2FFFE}1.19\%      &     \cellcolor[HTML]{E2FFFE}0.9211     &    \cellcolor[HTML]{E2FFFE}1.19\%       \\
\hline
\end{tabular}
}
\end{table*}

\begin{figure*}[!t]
\centering
\includegraphics[width=0.93\linewidth]{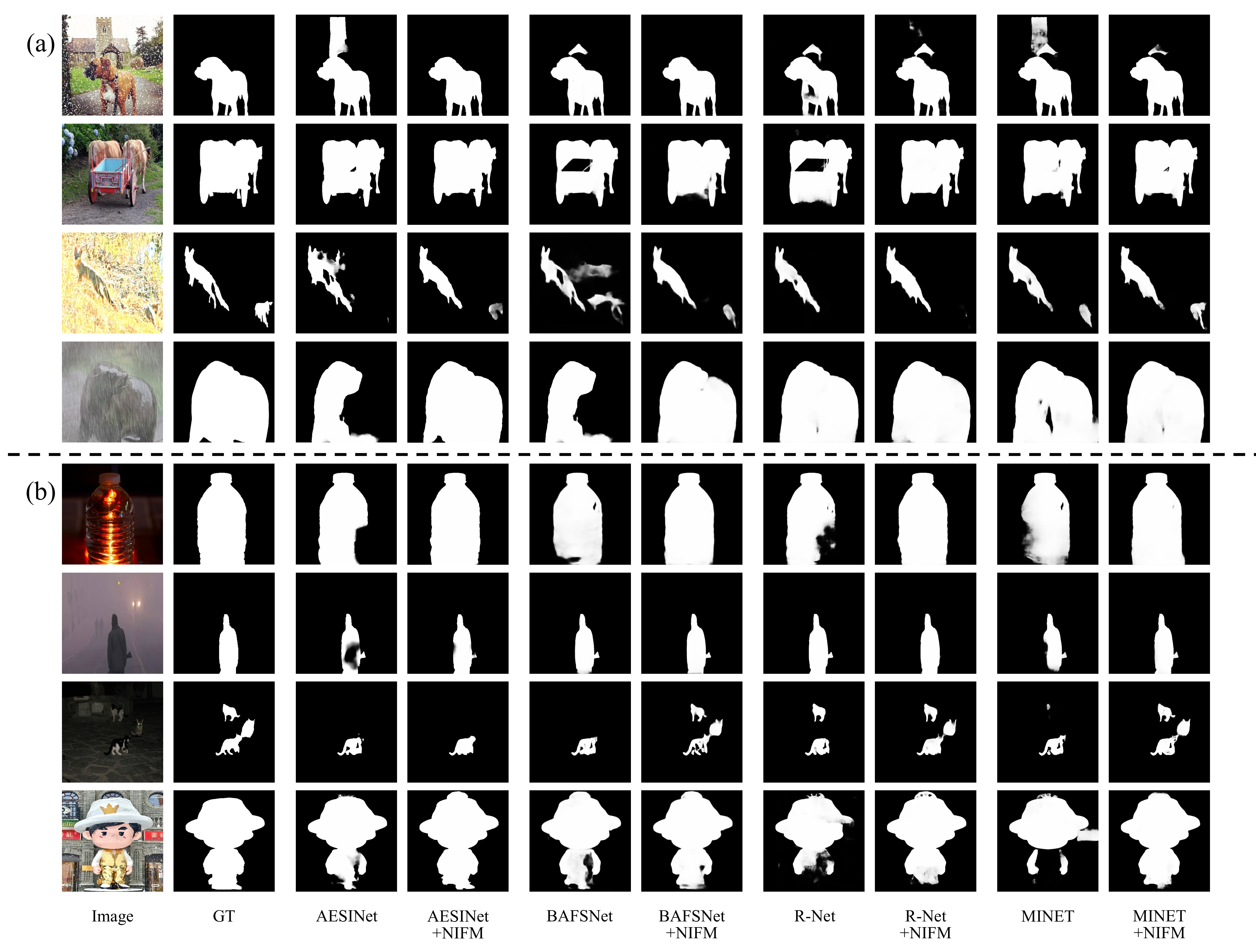}
\caption{Visual comparison under different decoder settings. (a) Visualization results on the synthesized test set. (b) Visualization results on the real test set.}
\label{visual_different_decoders}
\end{figure*}  

\begin{figure*}[!t]
\centering
\includegraphics[width=1.0\linewidth]{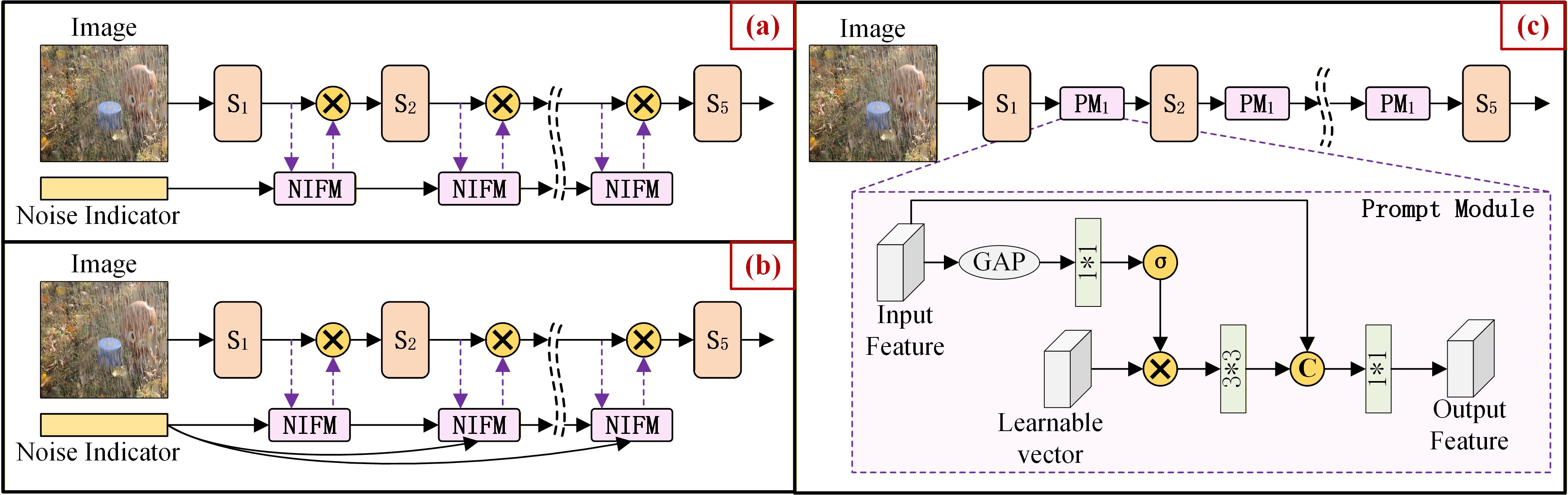}
\caption{Two schemes for integrating the SOD framework with NIFM.}
\label{NIFM variants}
\end{figure*}  

As presented in Table~\ref{results_100}, we report the results of models integrated with seven decoders (\textit{i.e.}, F3Net~\cite{wei2020f3net}, R-Net~\cite{wang2025r}, MINET~\cite{pang2020multi}, AESINet~\cite{zeng2023adaptive}, BAFSNet~\cite{gu2023orsi}, DC-Net~\cite{zhu2025dc} and EDN~\cite{wu2022edn}) on both synthesized and real test set of WXSOD.  Herein, "NIFM-\xmark" denotes models adopting the conventional ResNet-50 encoder, while "NIFM-\cmark" indicates models equipped with the specific encoder~(integrated with the noise indicator fusion module, NIFM).
Quantitative analysis reveals that after replacing the conventional encoder with our specific encoder, the average MAE across all seven models is reduced by 7.12\% on the synthesized test set and by 17.67\% on the real test set, respectively. 
Beyond MAE, consistent performance enhancements are also observed across other evaluation metrics, with greater improvement on the real test set compared to the synthesized one. This discrepancy highlights the specific encoder’s stronger ability to mitigate real weather noise, which exhibits more complex and unstructured characteristics than synthetic noise.
Furthermore, Table~\ref{results_100} also summarizes the variations in the computational cost~(MACs, Params and FPS) for models before and after embedding the NIFM. 
It is evident that NIFM is a lightweight module, increasing parameters by less than 2M and adding only 0.01G MACs.
Collectively, these results confirm that the proposed NIFM-enhanced specific encoder effectively boosts model performance with negligible computational cost, which is a core advantage for algorithm deployment.

We have provided additional experimental results for four models in Table~\ref{results_100}. Taking F3Net as an example, after injecting NIFM, the model performance using F3Net encoder surpasses TCGNet and A3Net, which also proves the effectiveness of our framework.

For comprehensive quantitative evaluation, Fig.~\ref{PRcurve} presents Precision-Recall~(PR) and F-measure curves, which compare the performance of models formed by pairing two types of encoders (the conventional encoder and our NIFM-enhanced encoder) with seven distinct decoders. A consistent trend can be observed across all decoder configurations: under the same decoder setup, the PR curve of the model equipped with the NIFM-enhanced encoder lies closer to the top-right corner, and the area below the F-measure curve is also larger than that of the model with the conventional encoder. These experiments results prove the effectiveness and superiority of our proposed NIFM-enhanced encoder.

To further evaluate the stability of the proposed framework across various dataset scales, all seven models are retrained using two reduced-scale training subsets of WXSOD, \textit{i.e}, 50\% Training set and 30\% Training set. 
The results are presented in Table~\ref{results5030}.
First, an intuitive trend can be observed: reducing the volume of training data leads to a corresponding decline in average performance across all seven models.
For instance, when the training set is scaled down from 100\% (original) to 50\% and 30\%, the average MAE on the real test set degrades from 0.0197 to 0.0228 and 0.0259, respectively.
Notably, even amid this performance degradation caused by reduced training data, the incorporation of our noise indicator fusion module (NIFM) consistently enhances segmentation accuracy across both the synthetic and real-world test subsets, regardless of the training set scale (50\% or 30\%). This consistent improvement underscores the NIFM’s ability to efficiently leverage available training data to learn noise-robust cues, mitigating the adverse effects of data scarcity. These findings confirm the feasibility and practicality of our proposed framework, especially in data-constrained real-world applications.


\subsubsection{ Qualitative Comparison}
To intuitively demonstrate the effectiveness of the proposed framework, specifically, the specialized encoder integrated with NIFM, we conducted a visual comparison using samples under four different weather scenarios from both the synthesized and real test sets.
Fig.~\ref{visual_different_decoders} presents the saliency prediction results of models configured with either the conventional encoder~(baseline) or the specific encoder (denoted as “+NIFM”), paired with four decoders~(AESINet, BAFSNet, Rnet, and F3Net).
It can be observed that the integration of NIFM yields more complete prediction results, demonstrating inherent robustness against weather-induced noise. 
For example, in the first row of Fig.~\ref{visual_different_decoders}(b), the contour of the bottle is blurred by low light. 
As a result, the model with the conventional encoder can only detect partial regions of the bottle with relatively high contrast. In contrast, the model integrated with NIFM produces a more complete detection.

\begin{figure*}[!t]
\centering
\includegraphics[width=1.0\linewidth]{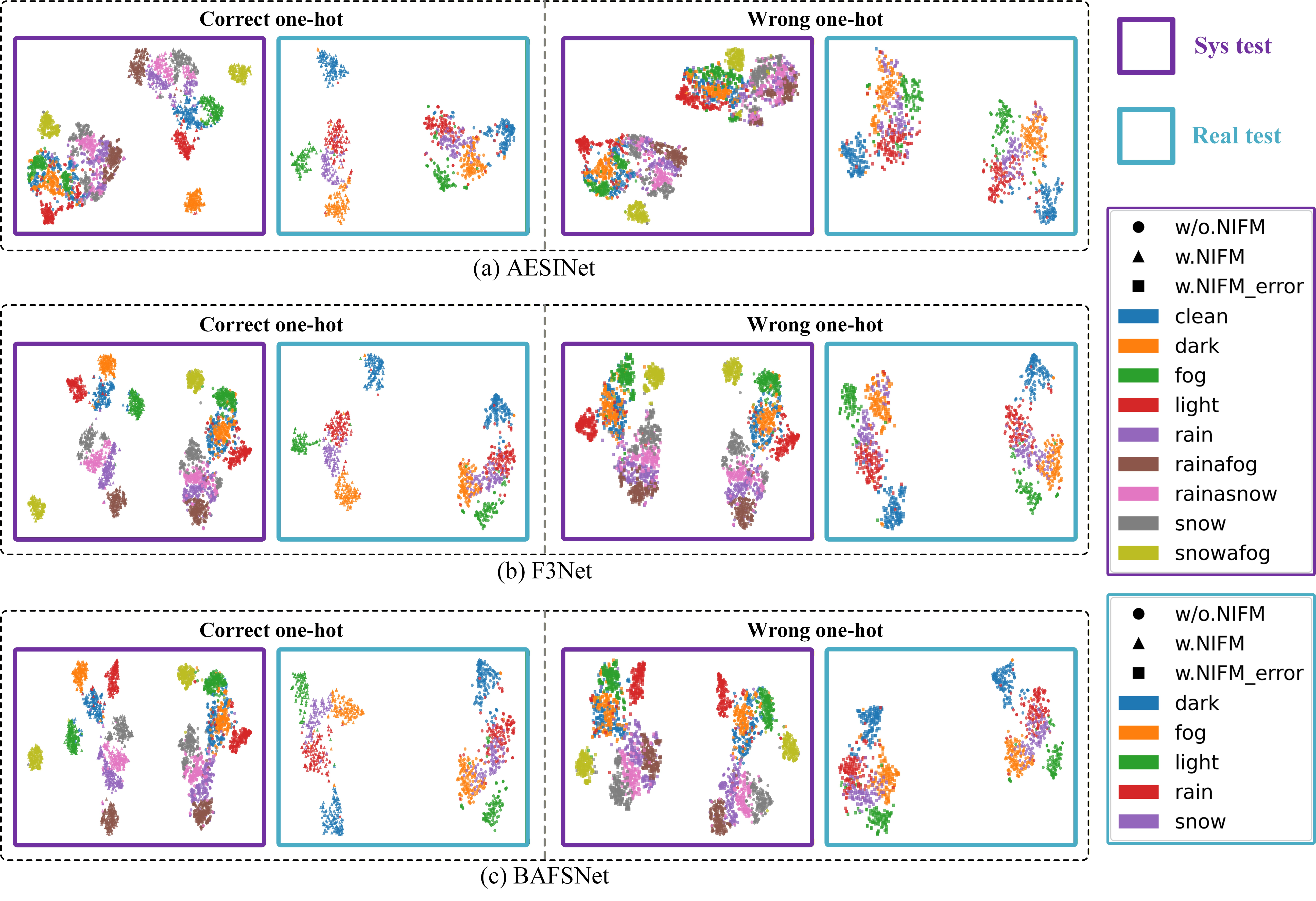}
\caption{Visualization of output features from different models via t-SNE. Purple boxes denote results from the synthetic test set while blue boxes denote those from the real test set. Circular features are derived from models without NIFM, square features from models with incorrectly configured one-hot encoding, and triangular features represent complete models with correctly configured one-hot encoding.}
\label{t-sne}
\end{figure*}

To verify the regulatory effect of NIFM on feature distributions, we visualized the feature distributions of various models using the t-SNE algorithm as shown in Fig.~\ref{t-sne}. 
Experiments under three decoder settings demonstrate that models incorporating NIFM with correct one-hot vectors achieve more distinct feature separation according to noise types on both real and synthetic test datasets, where features from different noise categories are well-separated.
In contrast, when the one-hot vectors in NIFM are incorrect (\textit{i.e.}, mismatched with the image noise categories), the feature distributions generated by NIFM-based models are almost indistinguishable from those of models without NIFM, and the feature distributions of the two types of models exhibit a symmetric structure. 
This fully validates the effectiveness of using noise indicators to distinguish image noise types and using NIFM to fuse semantic features with noise indicators.

\subsection{Ablation Studies}\label{Ablation Studies}

\begin{table*}[!t]
\renewcommand{\arraystretch}{1.10}
\centering
\caption{Comparison of Model Performance Using Two Distinct Integration strategies for NIFM. (a) denotes recursive fusion, while (b) represents hybrid fusion.}
\label{tab: two NIFM}
\resizebox{1.0\linewidth}{!}{
\begin{tabular}{c|c|c|cc|cc|cc|cc|cc|cc|cc|cc}
\hline
\multirow{2}{*}{Test set}&
\multirow{2}{*}{Decoders} &
\multirow{2}{*}{NIFM} &
\multicolumn{2}{c|}{${MAE} \downarrow$} &
\multicolumn{2}{c|}{${S} \uparrow$} &
\multicolumn{2}{c|}{${F}_{\beta}^{adp} \uparrow$} &
\multicolumn{2}{c|}{${F}_{\beta}^{mean} \uparrow$} &
\multicolumn{2}{c|}{${F}_{\beta}^{max} \uparrow$} &
\multicolumn{2}{c|}{${E}_{\xi}^{adp} \uparrow$} &
\multicolumn{2}{c|}{${E}_{\xi}^{mean} \uparrow$} &
\multicolumn{2}{c}{${E}_{\xi}^{max} \uparrow$} \\
& &  & val & $\Delta$ & val & $\Delta$ & val & $\Delta$ & val & $\Delta$ & val & $\Delta$ & val & $\Delta$ & val & $\Delta$ & val & $\Delta$ \\ \hline
\multirow{16}{*}{Sys}    & \multirow{3}{*}{F3Net}    &   \cellcolor[HTML]{DCF7FF}(a)  &   \cellcolor[HTML]{DCF7FF}0.0307        &   \cellcolor[HTML]{DCF7FF}13.03\%  &  \cellcolor[HTML]{DCF7FF}0.8845    &  \cellcolor[HTML]{DCF7FF}1.25\%   &  \cellcolor[HTML]{DCF7FF}0.8322     & \cellcolor[HTML]{DCF7FF}2.60\%    &   \cellcolor[HTML]{DCF7FF}0.8411    &    \cellcolor[HTML]{DCF7FF}2.23\%   &   \cellcolor[HTML]{DCF7FF}0.8587    &  \cellcolor[HTML]{DCF7FF}1.82\%     &  \cellcolor[HTML]{DCF7FF}0.9239   &   \cellcolor[HTML]{DCF7FF}1.15\%  &  \cellcolor[HTML]{DCF7FF}0.9200     &   \cellcolor[HTML]{DCF7FF}1.19\%    &    \cellcolor[HTML]{DCF7FF}0.9309       &  \cellcolor[HTML]{DCF7FF}1.05\%   \\
\hline
\end{tabular}
}
\end{table*}

%

\subsubsection{Impact of Different Integration Schemes of NIFMs on Model Performance}
To deeply analyze the effectiveness of integration modes between the NIFM and the encoder, we construct two variants of NIFM integration strategies, as illustrated in Fig.~\ref{NIFM variants}. Fig.~\ref{NIFM variants}(a) presents the recursive integration strategy, where the output weight of the preceding NIFM is taken as one of the input features for the subsequent NIFM. 
Fig.~\ref{NIFM variants}(b) depicts the hybrid integration strategy, which feeds the output weight of the previous NIFM as auxiliary features into the next NIFM based on the default integration mode. 
Meanwhile, we replace NIFM with a learnable prompt block~\cite{potlapalli2023promptir} for training and testing. 
This alternative approach is also a popular learning scheme for noise type identification, as illustrated in Fig.~\ref{NIFM variants}(c).
Subsequently, we retrained 21 models (combinations of 3 variant encoders and 7 decoders) using 100\% of the training set, and the experimental results are shown in Table~\ref{tab: two NIFM}. 
It should be noted that the calculation method of the improvement ratio is the same as that in Table~\ref{results_100}.
Due to space limitations, the results of Table~\ref{results_100} are not duplicated in Table~\ref{tab: two NIFM}.

As can be clearly observed from Table~\ref{tab: two NIFM}, across multiple groups of experiments on both the Synthesized test set and the Real test set, the recursive integration exhibits superior performance in most metrics. For instance, on the Synthesized test set, the average MAE of models under recursive integration is 0.0322 (+4.93\%), while that under hybrid integration is 0.0326 (3.88\%). In terms of metrics such as $S$ and $F_\beta^{adp}$, the average improvement amplitude of recursive integration is also more significant. On the Real test set, the average MAE of recursive integration is 0.0205 (+14.21\%), and that of hybrid integration is 0.0231 (+3.29\%). This indicates that compared with the hybrid integration strategy, the recursive integration strategy can transfer and fuse noise indication information more efficiently.

However, the performance gains generated by these two variants are both lower than that of the default integration strategy. Taking the average MAE as an example, the gain ratios of the default integration strategy on the synthetic test set and the real test set are 7.27\% and 17.67\%, respectively, which far exceed those of the recursive integration strategy (4.93\% and 14.21\%) and the hybrid integration strategy (3.88\% and 3.28\%). The aforementioned experimental results verify the crucial role of reasonably designing the integration mode of NIFM in improving the segmentation performance of the model under complex weather conditions.

Furthermore, models incorporating the prompt block exhibit decreased average performance on both synthesized and Real test set. We attribute this phenomenon primarily to the fact that optimizing the learnable prompt relies on sufficiently large datasets. Massive data is crucial for boosting model learning of noise specific representations.


\subsubsection{The Effectiveness of NIFM Meeting Various Backbone Networks}
To evaluate the compatibility of NIFM with different backbone networks, we replace the ResNet-50 in the framework with MobileNet~\cite{howard2017mobilenets} and MobileViT~\cite{mehta2021mobilevit}. 
As presented in Table~\ref{tab: different backbones}, on the real test set, the introduction of NIFM into the MobileNet and MobileViT backbones improved the average MAE across seven models by 14.78\% and 4.96\%, respectively, demonstrating the compatibility of NIFM with diverse backbone networks. 
On the other hand, the performance gains brought by NIFM are more pronounced in ResNet and MobileNet settings compared to MobileViT.
This phenomenon can be attributed to the superior feature extraction capability of Transformer-based architectures, which partially diminishes the contribution of noise-type guidance mechanisms.

\begin{table*}[!t]
\renewcommand{\arraystretch}{1.10}
\centering
\caption{Comparison of model performance using different backbones, namely MobileNet and MobileViT.}
\label{tab: different backbones}
\resizebox{1.0\linewidth}{!}{
\begin{tabular}{c|c|c|cccccccc|cccccccc}
\hline
&                           &                        & \multicolumn{8}{c|}{\cellcolor[HTML]{DCF7FF}MobileNet}           & \multicolumn{8}{c}{\cellcolor[HTML]{FFFFE0}MobileViT}     \\ \cline{4-19} 
&           &          & \multicolumn{2}{c|}{$MAE\downarrow$}       & \multicolumn{2}{c|}{$S\uparrow$}    & \multicolumn{2}{c|}{$F_{\beta}^{mean}\uparrow$}    & \multicolumn{2}{c|}{$E_{\xi}^{mean}\uparrow$}   & \multicolumn{2}{c|}{$MAE\downarrow$}     & \multicolumn{2}{c|}{$S\uparrow$}     & \multicolumn{2}{c|}{$F_{\beta}^{mean}\uparrow$}   & \multicolumn{2}{c}{$E_{\xi}^{mean}\uparrow$}    \\
\multirow{-3}{*}{Test set} & \multirow{-3}{*}{Decoder} & \multirow{-3}{*}{NIFM} & val               & \multicolumn{1}{c|}{$\Delta$}       & val         & \multicolumn{1}{c|}{$\Delta$}           & val      & \multicolumn{1}{c|}{$\Delta$}      & val        & $\Delta$        & val         & \multicolumn{1}{c|}{$\Delta$}     & val          & \multicolumn{1}{c|}{$\Delta$}       & val        & \multicolumn{1}{c|}{$\Delta$}     & val       & $\Delta$    \\
\hline
&        & \xmark  & 0.0583 & \multicolumn{1}{c|}{}         & 0.8011 & \multicolumn{1}{c|}{}      & 0.7216 & \multicolumn{1}{c|}{}                         & 0.8455 &            & 0.0447 & \multicolumn{1}{c|}{}                         & 0.8445 & \multicolumn{1}{c|}{}               & 0.7839 & \multicolumn{1}{c|}{}             & 0.8874 &               \\
& \multirow{-2}{*}{F3Net}   & \cmark  & \cellcolor[HTML]{DCF7FF}0.0579 & \multicolumn{1}{l|}{\cellcolor[HTML]{DCF7FF}0.69\%} & \cellcolor[HTML]{DCF7FF}0.8035 & \multicolumn{1}{l|}{\cellcolor[HTML]{DCF7FF}0.30\%} & \cellcolor[HTML]{DCF7FF}0.7261 & \multicolumn{1}{l|}{\cellcolor[HTML]{DCF7FF}0.62\%} & \cellcolor[HTML]{DCF7FF}0.8462 & \cellcolor[HTML]{DCF7FF}0.08\% & \cellcolor[HTML]{FFFFE0}0.0424 & \multicolumn{1}{l|}{\cellcolor[HTML]{FFFFE0}5.15\%} & \cellcolor[HTML]{FFFFE0}0.8498 & \multicolumn{1}{l|}{\cellcolor[HTML]{FFFFE0}0.62\%} & \cellcolor[HTML]{FFFFE0}0.7935 & \multicolumn{1}{l|}{\cellcolor[HTML]{FFFFE0}1.23\%} & \cellcolor[HTML]{FFFFE0}0.8943 & \cellcolor[HTML]{FFFFE0}0.78\% \\
\cdashline{2-19}[5pt/3pt]
&                           & \xmark  & 0.0368 & \multicolumn{1}{l|}{}                         & 0.8679 & \multicolumn{1}{l|}{}                         & 0.8154 & \multicolumn{1}{l|}{}                         & 0.9058 &                          & 0.0462 & \multicolumn{1}{l|}{}                         & 0.8350 & \multicolumn{1}{l|}{}                         & 0.7745 & \multicolumn{1}{l|}{}                         & 0.8751 &                          \\
& \multirow{-2}{*}{MINET}   & \cmark  & \cellcolor[HTML]{DCF7FF}0.0333 & \multicolumn{1}{l|}{\cellcolor[HTML]{DCF7FF}9.51\%} & \cellcolor[HTML]{DCF7FF}0.8748 & \multicolumn{1}{l|}{\cellcolor[HTML]{DCF7FF}0.79\%} & \cellcolor[HTML]{DCF7FF}0.8282 & \multicolumn{1}{l|}{\cellcolor[HTML]{DCF7FF}1.57\%} & \cellcolor[HTML]{DCF7FF}0.9119 & \cellcolor[HTML]{DCF7FF}0.67\% & \cellcolor[HTML]{FFFFE0}0.0466 & \multicolumn{1}{l|}{\cellcolor[HTML]{FFFFE0}-0.87\%} & \cellcolor[HTML]{FFFFE0}0.8370 & \multicolumn{1}{l|}{\cellcolor[HTML]{FFFFE0}0.24\%} & \cellcolor[HTML]{FFFFE0}0.7769 & \multicolumn{1}{l|}{\cellcolor[HTML]{FFFFE0}0.30\%} & \cellcolor[HTML]{FFFFE0}0.8799 & \cellcolor[HTML]{FFFFE0}0.55\% \\\cdashline{2-19}[5pt/3pt]
&                           & \xmark  & 0.0376 & \multicolumn{1}{l|}{}                         & 0.8637 & \multicolumn{1}{l|}{}                         & 0.8039 & \multicolumn{1}{l|}{}                         & 0.8958 &                          & 0.0475 & \multicolumn{1}{l|}{}                         & 0.8379 & \multicolumn{1}{l|}{}                         & 0.7647 & \multicolumn{1}{l|}{}                         & 0.8688 &                          \\
& \multirow{-2}{*}{EDN}     & \cmark  & \cellcolor[HTML]{DCF7FF}0.0341 & \multicolumn{1}{l|}{\cellcolor[HTML]{DCF7FF}9.31\%} & \cellcolor[HTML]{DCF7FF}0.8710 & \multicolumn{1}{l|}{\cellcolor[HTML]{DCF7FF}0.84\%} & \cellcolor[HTML]{DCF7FF}0.8160 & \multicolumn{1}{l|}{\cellcolor[HTML]{DCF7FF}1.51\%} & \cellcolor[HTML]{DCF7FF}0.9052 & \cellcolor[HTML]{DCF7FF}1.05\% & \cellcolor[HTML]{FFFFE0}0.0457 & \multicolumn{1}{l|}{\cellcolor[HTML]{FFFFE0}3.79\%} & \cellcolor[HTML]{FFFFE0}0.8410 & \multicolumn{1}{l|}{\cellcolor[HTML]{FFFFE0}0.36\%} & \cellcolor[HTML]{FFFFE0}0.7717 & \multicolumn{1}{l|}{\cellcolor[HTML]{FFFFE0}0.91\%} & \cellcolor[HTML]{FFFFE0}0.8716 & \cellcolor[HTML]{FFFFE0}0.33\% \\\cdashline{2-19}[5pt/3pt]
&                           & \xmark  & 0.0342 & \multicolumn{1}{l|}{}                         & 0.8526 & \multicolumn{1}{l|}{}                         & 0.8050 & \multicolumn{1}{l|}{}                         & 0.8957 &                          & 0.0478 & \multicolumn{1}{l|}{}                         & 0.8197 & \multicolumn{1}{l|}{}                         & 0.7526 & \multicolumn{1}{l|}{}                         & 0.8761 &                          \\
& \multirow{-2}{*}{AESINet} & \cmark  & \cellcolor[HTML]{DCF7FF}0.0354 & \multicolumn{1}{l|}{\cellcolor[HTML]{DCF7FF}-3.51\%} & \cellcolor[HTML]{DCF7FF}0.8604 & \multicolumn{1}{l|}{\cellcolor[HTML]{DCF7FF}0.91\%} & \cellcolor[HTML]{DCF7FF}0.8152 & \multicolumn{1}{l|}{\cellcolor[HTML]{DCF7FF}1.27\%} & \cellcolor[HTML]{DCF7FF}0.9049 & \cellcolor[HTML]{DCF7FF}1.03\% & \cellcolor[HTML]{FFFFE0}0.0484 & \multicolumn{1}{l|}{\cellcolor[HTML]{FFFFE0}-1.26\%} & \cellcolor[HTML]{FFFFE0}0.8219 & \multicolumn{1}{l|}{\cellcolor[HTML]{FFFFE0}0.27\%} & \cellcolor[HTML]{FFFFE0}0.7608 & \multicolumn{1}{l|}{\cellcolor[HTML]{FFFFE0}1.09\%} & \cellcolor[HTML]{FFFFE0}0.8738 & \cellcolor[HTML]{FFFFE0}-0.26\% \\\cdashline{2-19}[5pt/3pt]
&                           & \xmark  & 0.0315 & \multicolumn{1}{l|}{}                         & 0.8829 & \multicolumn{1}{l|}{}                         & 0.8358 & \multicolumn{1}{l|}{}                         & 0.9193 &                          & 0.0416 & \multicolumn{1}{l|}{}                         & 0.8453 & \multicolumn{1}{l|}{}                         & 0.7870 & \multicolumn{1}{l|}{}                         & 0.8869 &                          \\
& \multirow{-2}{*}{BAFSNet} & \cmark  & \cellcolor[HTML]{DCF7FF}0.0322 & \multicolumn{1}{l|}{\cellcolor[HTML]{DCF7FF}-2.22\%} & \cellcolor[HTML]{DCF7FF}0.8823 & \multicolumn{1}{l|}{\cellcolor[HTML]{DCF7FF}-0.07\%} & \cellcolor[HTML]{DCF7FF}0.8360 & \multicolumn{1}{l|}{\cellcolor[HTML]{DCF7FF}0.01\%} & \cellcolor[HTML]{DCF7FF}0.9177 & \cellcolor[HTML]{DCF7FF}-0.17\% & \cellcolor[HTML]{FFFFE0}0.0426 & \multicolumn{1}{l|}{\cellcolor[HTML]{FFFFE0}-2.40\%} & \cellcolor[HTML]{FFFFE0}0.8439 & \multicolumn{1}{l|}{\cellcolor[HTML]{FFFFE0}-0.17\%} & \cellcolor[HTML]{FFFFE0}0.7873 & \multicolumn{1}{l|}{\cellcolor[HTML]{FFFFE0}0.04\%} & \cellcolor[HTML]{FFFFE0}0.8885 & \cellcolor[HTML]{FFFFE0}0.18\% \\\cdashline{2-19}[5pt/3pt]
&                           & \xmark  & 0.0354 & \multicolumn{1}{l|}{}                         & 0.8682 & \multicolumn{1}{l|}{}                         & 0.8225 & \multicolumn{1}{l|}{}                         & 0.9066 &                          & 0.0388 & \multicolumn{1}{l|}{}                         & 0.8600 & \multicolumn{1}{l|}{}                         & 0.8036 & \multicolumn{1}{l|}{}                         & 0.8995 &                          \\
& \multirow{-2}{*}{DC-Net}  & \cmark  & \cellcolor[HTML]{DCF7FF}0.0331 & \multicolumn{1}{l|}{\cellcolor[HTML]{DCF7FF}6.50\%} & \cellcolor[HTML]{DCF7FF}0.8771 & \multicolumn{1}{l|}{\cellcolor[HTML]{DCF7FF}1.03\%} & \cellcolor[HTML]{DCF7FF}0.8276 & \multicolumn{1}{l|}{\cellcolor[HTML]{DCF7FF}0.61\%} & \cellcolor[HTML]{DCF7FF}0.9160 & \cellcolor[HTML]{DCF7FF}1.04\% & \cellcolor[HTML]{FFFFE0}0.0414 & \multicolumn{1}{l|}{\cellcolor[HTML]{FFFFE0}-6.70\%} & \cellcolor[HTML]{FFFFE0}0.8462 & \multicolumn{1}{l|}{\cellcolor[HTML]{FFFFE0}-1.60\%} & \cellcolor[HTML]{FFFFE0}0.7884 & \multicolumn{1}{l|}{\cellcolor[HTML]{FFFFE0}-1.89\%} & \cellcolor[HTML]{FFFFE0}0.8899 & \cellcolor[HTML]{FFFFE0}-1.06\% \\\cdashline{2-19}[5pt/3pt]
&                           & \xmark  & 0.0509 & \multicolumn{1}{l|}{}                         & 0.8112 & \multicolumn{1}{l|}{}                         & 0.7469 & \multicolumn{1}{l|}{}                         & 0.8585 &                          & 0.0637 & \multicolumn{1}{l|}{}                         & 0.7643 & \multicolumn{1}{l|}{}                         & 0.6787 & \multicolumn{1}{l|}{}                         & 0.8023 &                          \\
& \multirow{-2}{*}{R-Net}   & \cmark  & \cellcolor[HTML]{DCF7FF}0.0506 & \multicolumn{1}{l|}{\cellcolor[HTML]{DCF7FF}0.59\%} & \cellcolor[HTML]{DCF7FF}0.8217 & \multicolumn{1}{l|}{\cellcolor[HTML]{DCF7FF}1.30\%} & \cellcolor[HTML]{DCF7FF}0.7593 & \multicolumn{1}{l|}{\cellcolor[HTML]{DCF7FF}1.66\%} & \cellcolor[HTML]{DCF7FF}0.8632 & \cellcolor[HTML]{DCF7FF}0.55\% & \cellcolor[HTML]{FFFFE0}0.0604 & \multicolumn{1}{l|}{\cellcolor[HTML]{FFFFE0}5.18\%} & \cellcolor[HTML]{FFFFE0}0.7904 & \multicolumn{1}{l|}{\cellcolor[HTML]{FFFFE0}3.42\%} & \cellcolor[HTML]{FFFFE0}0.7143 & \multicolumn{1}{l|}{\cellcolor[HTML]{FFFFE0}5.24\%} & \cellcolor[HTML]{FFFFE0}0.8340 & \cellcolor[HTML]{FFFFE0}3.96\% \\\cdashline{2-19}[5pt/3pt]
&                           & \xmark  & 0.0407 & \multicolumn{1}{l|}{}                         & 0.8497 & \multicolumn{1}{l|}{}                         & 0.7930 & \multicolumn{1}{l|}{}                         & 0.8896 &                          & 0.0479 & \multicolumn{1}{l|}{}                         & 0.8287 & \multicolumn{1}{l|}{}                         & 0.7611 & \multicolumn{1}{l|}{}                         & 0.8702 &                          \\
\multirow{-16}{*}{Sys}     & \multirow{-2}{*}{Average} & \cmark  & \cellcolor[HTML]{DCF7FF}0.0395 & \multicolumn{1}{l|}{\cellcolor[HTML]{DCF7FF}2.85\%} & \cellcolor[HTML]{DCF7FF}0.8558 & \multicolumn{1}{l|}{\cellcolor[HTML]{DCF7FF}0.73\%} & \cellcolor[HTML]{DCF7FF}0.8012 & \multicolumn{1}{l|}{\cellcolor[HTML]{DCF7FF}1.03\%} & \cellcolor[HTML]{DCF7FF}0.8950 & \cellcolor[HTML]{DCF7FF}0.61\% & \cellcolor[HTML]{FFFFE0}0.0486 & \multicolumn{1}{l|}{\cellcolor[HTML]{FFFFE0}-1.49\%} & \cellcolor[HTML]{FFFFE0}0.8275 & \multicolumn{1}{l|}{\cellcolor[HTML]{FFFFE0}-0.14\%} & \cellcolor[HTML]{FFFFE0}0.7612 & \multicolumn{1}{l|}{\cellcolor[HTML]{FFFFE0}0.02\%} & \cellcolor[HTML]{FFFFE0}0.8709 & \cellcolor[HTML]{FFFFE0}0.09\% \\ 
\hline
\hline
&                           & \xmark  & 0.0411 & \multicolumn{1}{l|}{}                         & 0.8344 & \multicolumn{1}{l|}{}                         & 0.7464 & \multicolumn{1}{l|}{}                         & 0.8623 &                          & 0.0324 & \multicolumn{1}{l|}{}                         & 0.8754 & \multicolumn{1}{l|}{}                         & 0.8115 & \multicolumn{1}{l|}{}                         & 0.9063 &                          \\
& \multirow{-2}{*}{F3Net}   & \cmark  & \cellcolor[HTML]{DCF7FF}0.0373 & \multicolumn{1}{l|}{\cellcolor[HTML]{DCF7FF}9.25\%} & \cellcolor[HTML]{DCF7FF}0.8388 & \multicolumn{1}{l|}{\cellcolor[HTML]{DCF7FF}0.53\%} & \cellcolor[HTML]{DCF7FF}0.7527 & \multicolumn{1}{l|}{\cellcolor[HTML]{DCF7FF}0.84\%} & \cellcolor[HTML]{DCF7FF}0.8587 & \cellcolor[HTML]{DCF7FF}-0.41\% & \cellcolor[HTML]{FFFFE0}0.0272 & \multicolumn{1}{l|}{\cellcolor[HTML]{FFFFE0}16.05\%} & \cellcolor[HTML]{FFFFE0}0.8884 & \multicolumn{1}{l|}{\cellcolor[HTML]{FFFFE0}1.48\%} & \cellcolor[HTML]{FFFFE0}0.8332 & \multicolumn{1}{l|}{\cellcolor[HTML]{FFFFE0}2.67\%} & \cellcolor[HTML]{FFFFE0}0.9166 & \cellcolor[HTML]{FFFFE0}1.13\% \\\cdashline{2-19}[5pt/3pt]
&                           & \xmark  & 0.0264 & \multicolumn{1}{l|}{}                         & 0.8972 & \multicolumn{1}{l|}{}                         & 0.8455 & \multicolumn{1}{l|}{}                         & 0.9226 &                          & 0.0296 & \multicolumn{1}{l|}{}                         & 0.8787 & \multicolumn{1}{l|}{}                         & 0.8200 & \multicolumn{1}{l|}{}                         & 0.9051 &                          \\
& \multirow{-2}{*}{MINET}   & \cmark  & \cellcolor[HTML]{DCF7FF}0.0227 & \multicolumn{1}{l|}{\cellcolor[HTML]{DCF7FF}14.02\%} & \cellcolor[HTML]{DCF7FF}0.9037 & \multicolumn{1}{l|}{\cellcolor[HTML]{DCF7FF}0.73\%} & \cellcolor[HTML]{DCF7FF}0.8501 & \multicolumn{1}{l|}{\cellcolor[HTML]{DCF7FF}0.54\%} & \cellcolor[HTML]{DCF7FF}0.9247 & \cellcolor[HTML]{DCF7FF}0.23\% & \cellcolor[HTML]{FFFFE0}0.0279 & \multicolumn{1}{l|}{\cellcolor[HTML]{FFFFE0}5.74\%} & \cellcolor[HTML]{FFFFE0}0.8846 & \multicolumn{1}{l|}{\cellcolor[HTML]{FFFFE0}0.68\%} & \cellcolor[HTML]{FFFFE0}0.8190 & \multicolumn{1}{l|}{\cellcolor[HTML]{FFFFE0}-0.13\%} & \cellcolor[HTML]{FFFFE0}0.9166 & \cellcolor[HTML]{FFFFE0}1.27\% \\\cdashline{2-19}[5pt/3pt]
&                           & \xmark  & 0.0357 & \multicolumn{1}{l|}{}                         & 0.8810 & \multicolumn{1}{l|}{}                         & 0.8108 & \multicolumn{1}{l|}{}                         & 0.8972 &                          & 0.0327 & \multicolumn{1}{l|}{}                         & 0.8680 & \multicolumn{1}{l|}{}                         & 0.7954 & \multicolumn{1}{l|}{}                         & 0.8922 &                          \\
& \multirow{-2}{*}{EDN}     & \cmark  & \cellcolor[HTML]{DCF7FF}0.0229 & \multicolumn{1}{l|}{\cellcolor[HTML]{DCF7FF}35.85\%} & \cellcolor[HTML]{DCF7FF}0.9029 & \multicolumn{1}{l|}{\cellcolor[HTML]{DCF7FF}2.48\%} & \cellcolor[HTML]{DCF7FF}0.8382 & \multicolumn{1}{l|}{\cellcolor[HTML]{DCF7FF}3.38\%} & \cellcolor[HTML]{DCF7FF}0.9207 & \cellcolor[HTML]{DCF7FF}2.63\% & \cellcolor[HTML]{FFFFE0}0.0291 & \multicolumn{1}{l|}{\cellcolor[HTML]{FFFFE0}11.01\%} & \cellcolor[HTML]{FFFFE0}0.8749 & \multicolumn{1}{l|}{\cellcolor[HTML]{FFFFE0}0.80\%} & \cellcolor[HTML]{FFFFE0}0.8103 & \multicolumn{1}{l|}{\cellcolor[HTML]{FFFFE0}1.86\%} & \cellcolor[HTML]{FFFFE0}0.9015 & \cellcolor[HTML]{FFFFE0}1.05\% \\\cdashline{2-19}[5pt/3pt]
&                           & \xmark  & 0.0273 & \multicolumn{1}{l|}{}                         & 0.8753 & \multicolumn{1}{l|}{}                         & 0.8170 & \multicolumn{1}{l|}{}                         & 0.9051 &                          & 0.0272 & \multicolumn{1}{l|}{}                         & 0.8683 & \multicolumn{1}{l|}{}                         & 0.7997 & \multicolumn{1}{l|}{}                         & 0.9120 &                          \\
& \multirow{-2}{*}{AESINet} & \cmark  & \cellcolor[HTML]{DCF7FF}0.0240 & \multicolumn{1}{l|}{\cellcolor[HTML]{DCF7FF}12.09\%} & \cellcolor[HTML]{DCF7FF}0.8980 & \multicolumn{1}{l|}{\cellcolor[HTML]{DCF7FF}2.58\%} & \cellcolor[HTML]{DCF7FF}0.8457 & \multicolumn{1}{l|}{\cellcolor[HTML]{DCF7FF}3.52\%} & \cellcolor[HTML]{DCF7FF}0.9303 & \cellcolor[HTML]{DCF7FF}2.78\% & \cellcolor[HTML]{FFFFE0}0.0267 & \multicolumn{1}{l|}{\cellcolor[HTML]{FFFFE0}1.84\%} & \cellcolor[HTML]{FFFFE0}0.8753 & \multicolumn{1}{l|}{\cellcolor[HTML]{FFFFE0}0.81\%} & \cellcolor[HTML]{FFFFE0}0.8166 & \multicolumn{1}{l|}{\cellcolor[HTML]{FFFFE0}2.11\%} & \cellcolor[HTML]{FFFFE0}0.9140 & \cellcolor[HTML]{FFFFE0}0.21\% \\\cdashline{2-19}[5pt/3pt]
&                           & \xmark  & 0.0242 & \multicolumn{1}{l|}{}                         & 0.8983 & \multicolumn{1}{l|}{}                         & 0.8366 & \multicolumn{1}{l|}{}                         & 0.9217 &                          & 0.0275 & \multicolumn{1}{l|}{}                         & 0.8795 & \multicolumn{1}{l|}{}                         & 0.8201 & \multicolumn{1}{l|}{}                         & 0.9174 &                          \\
& \multirow{-2}{*}{BAFSNet} & \cmark  & \cellcolor[HTML]{DCF7FF}0.0213 & \multicolumn{1}{l|}{\cellcolor[HTML]{DCF7FF}11.98\%} & \cellcolor[HTML]{DCF7FF}0.9076 & \multicolumn{1}{l|}{\cellcolor[HTML]{DCF7FF}1.03\%} & \cellcolor[HTML]{DCF7FF}0.8541 & \multicolumn{1}{l|}{\cellcolor[HTML]{DCF7FF}2.09\%} & \cellcolor[HTML]{DCF7FF}0.9316 & \cellcolor[HTML]{DCF7FF}1.07\% & \cellcolor[HTML]{FFFFE0}0.0251 & \multicolumn{1}{l|}{\cellcolor[HTML]{FFFFE0}8.73\%} & \cellcolor[HTML]{FFFFE0}0.8910 & \multicolumn{1}{l|}{\cellcolor[HTML]{FFFFE0}1.30\%} & \cellcolor[HTML]{FFFFE0}0.8352 & \multicolumn{1}{l|}{\cellcolor[HTML]{FFFFE0}1.84\%} & \cellcolor[HTML]{FFFFE0}0.9226 & \cellcolor[HTML]{FFFFE0}0.57\% \\\cdashline{2-19}[5pt/3pt]
&                           & \xmark  & 0.0264 & \multicolumn{1}{l|}{}                         & 0.8954 & \multicolumn{1}{l|}{}                         & 0.8401 & \multicolumn{1}{l|}{}                         & 0.9246 &                          & 0.0250 & \multicolumn{1}{l|}{}                         & 0.8860 & \multicolumn{1}{l|}{}                         & 0.8263 & \multicolumn{1}{l|}{}                         & 0.9221 &                          \\
& \multirow{-2}{*}{DC-Net}  & \cmark  & \cellcolor[HTML]{DCF7FF}0.0225 & \multicolumn{1}{l|}{\cellcolor[HTML]{DCF7FF}14.77\%} & \cellcolor[HTML]{DCF7FF}0.9054 & \multicolumn{1}{l|}{\cellcolor[HTML]{DCF7FF}1.11\%} & \cellcolor[HTML]{DCF7FF}0.8426 & \multicolumn{1}{l|}{\cellcolor[HTML]{DCF7FF}0.30\%} & \cellcolor[HTML]{DCF7FF}0.9262 & \cellcolor[HTML]{DCF7FF}0.18\% & \cellcolor[HTML]{FFFFE0}0.0256 & \multicolumn{1}{l|}{\cellcolor[HTML]{FFFFE0}-2.40\%} & \cellcolor[HTML]{FFFFE0}0.8828 & \multicolumn{1}{l|}{\cellcolor[HTML]{FFFFE0}-0.36\%} & \cellcolor[HTML]{FFFFE0}0.8294 & \multicolumn{1}{l|}{\cellcolor[HTML]{FFFFE0}0.38\%} & \cellcolor[HTML]{FFFFE0}0.9183 & \cellcolor[HTML]{FFFFE0}-0.41\% \\\cdashline{2-19}[5pt/3pt]
&                           & \xmark  & 0.0334 & \multicolumn{1}{l|}{}                         & 0.8662 & \multicolumn{1}{l|}{}                         & 0.7998 & \multicolumn{1}{l|}{}                         & 0.8947 &                          & 0.0438 & \multicolumn{1}{l|}{}                         & 0.8199 & \multicolumn{1}{l|}{}                         & 0.7431 & \multicolumn{1}{l|}{}                         & 0.8470 &                          \\
& \multirow{-2}{*}{R-Net}   & \cmark  & \cellcolor[HTML]{DCF7FF}0.0321 & \multicolumn{1}{l|}{\cellcolor[HTML]{DCF7FF}3.89\%} & \cellcolor[HTML]{DCF7FF}0.8651 & \multicolumn{1}{l|}{\cellcolor[HTML]{DCF7FF}-0.13\%} & \cellcolor[HTML]{DCF7FF}0.7965 & \multicolumn{1}{l|}{\cellcolor[HTML]{DCF7FF}-0.40\%} & \cellcolor[HTML]{DCF7FF}0.8880 & \cellcolor[HTML]{DCF7FF}-0.75\% & \cellcolor[HTML]{FFFFE0}0.0337 & \multicolumn{1}{l|}{\cellcolor[HTML]{FFFFE0}23.06\%} & \cellcolor[HTML]{FFFFE0}0.8513 & \multicolumn{1}{l|}{\cellcolor[HTML]{FFFFE0}3.83\%} & \cellcolor[HTML]{FFFFE0}0.7837 & \multicolumn{1}{l|}{\cellcolor[HTML]{FFFFE0}5.47\%} & \cellcolor[HTML]{FFFFE0}0.8847 & \cellcolor[HTML]{FFFFE0}4.46\% \\\cdashline{2-19}[5pt/3pt]
&                           & \xmark  & 0.0306 & \multicolumn{1}{l|}{}                         & 0.8783 & \multicolumn{1}{l|}{}                         & 0.8138 & \multicolumn{1}{l|}{}                         & 0.9040 &                          & 0.0322 & \multicolumn{1}{l|}{}                         & 0.8644 & \multicolumn{1}{l|}{}                         & 0.7956 & \multicolumn{1}{l|}{}                         & 0.8969 &                          \\
\multirow{-16}{*}{Real}    & \multirow{-2}{*}{Average} & \cmark  & \cellcolor[HTML]{DCF7FF}0.0261 & \multicolumn{1}{l|}{\cellcolor[HTML]{DCF7FF}14.78\%} & \cellcolor[HTML]{DCF7FF}0.8888 & \multicolumn{1}{l|}{\cellcolor[HTML]{DCF7FF}1.20\%} & \cellcolor[HTML]{DCF7FF}0.8257 & \multicolumn{1}{l|}{\cellcolor[HTML]{DCF7FF}1.47\%} & \cellcolor[HTML]{DCF7FF}0.9115 & \cellcolor[HTML]{DCF7FF}0.82\% & \cellcolor[HTML]{FFFFE0}0.0306 & \multicolumn{1}{l|}{\cellcolor[HTML]{FFFFE0}4.96\%} & \cellcolor[HTML]{FFFFE0}0.8667 & \multicolumn{1}{l|}{\cellcolor[HTML]{FFFFE0}0.26\%} & \cellcolor[HTML]{FFFFE0}0.8001 & \multicolumn{1}{l|}{\cellcolor[HTML]{FFFFE0}0.56\%} & \cellcolor[HTML]{FFFFE0}0.8965 & \cellcolor[HTML]{FFFFE0}-0.03\% \\ \hline 
\end{tabular}
}
\end{table*}


\subsubsection{Impact of Training Data Diversity on Model Performance}

\begin{table*}[!t]
\renewcommand{\arraystretch}{1.0}
\centering
\caption{Quantitative Comparison Results of $MAE$, $S$, $F_{\beta}^{mean}$ and $E_{\xi}^{mean}$ on the WXSOD dataset. Here, "$\uparrow$" ($\downarrow$) means that the larger (smaller) the better. The "Sys" and "Real" denotes the synthesized and real test set, respectively. The best results are highlighted in \textbf{bold}.}
\label{tab:quantitative_comparison_subsets}
\resizebox{0.95\linewidth}{!}{
\begin{tabular}{c|c|c|cccccccccccc}
\hline
\multirow{2}{*}{Test set}  & \multirow{2}{*}{Method} & \multirow{2}{*}{Metrics}  & \multicolumn{11}{c}{Training subset}             \\\cline{4-14}
&      &   & Clean & Rain & Snow & Fog & Light & Dark & Rain\&Fog & Rain\&Snow & Snow\&Fog & Multi-weather & \begin{tabular}[c]{@{}c@{}}Multi-weather\\ (w.NIFM)\end{tabular} \\ \hline
\multirow{32}{*}{Sys}   & \multirow{4}{*}{F3Net}      & \cellcolor[HTML]{DCF7FF}$MAE\downarrow$ &    \cellcolor[HTML]{DCF7FF}0.0701 &   \cellcolor[HTML]{DCF7FF}0.0676 &  \cellcolor[HTML]{DCF7FF}0.0570    &   \cellcolor[HTML]{DCF7FF}0.0796  &   \cellcolor[HTML]{DCF7FF}0.0697    &    \cellcolor[HTML]{DCF7FF}0.0817  &    \cellcolor[HTML]{DCF7FF}0.0887   &     \cellcolor[HTML]{DCF7FF}0.0606   &     \cellcolor[HTML]{DCF7FF}0.1200    &    \cellcolor[HTML]{DCF7FF}0.0353   &     \cellcolor[HTML]{DCF7FF}0.0307   \\
&      & \cellcolor[HTML]{FFFFE0}$S\uparrow$ & \cellcolor[HTML]{FFFFE0}0.7396      &  \cellcolor[HTML]{FFFFE0}0.7429    &  \cellcolor[HTML]{FFFFE0}0.7914    &   \cellcolor[HTML]{FFFFE0}0.7124  &  \cellcolor[HTML]{FFFFE0}0.7337     &  \cellcolor[HTML]{FFFFE0}0.7208    & \cellcolor[HTML]{FFFFE0}0.6403    &  \cellcolor[HTML]{FFFFE0}0.7837   &  \cellcolor[HTML]{FFFFE0}0.6633    &  \cellcolor[HTML]{FFFFE0}0.8735  &   \cellcolor[HTML]{FFFFE0}0.8822   \\
&      &  \cellcolor[HTML]{FFEFEF}$F_{\beta}^{mean}\uparrow$ & \cellcolor[HTML]{FFEFEF}0.6146      &  \cellcolor[HTML]{FFEFEF}0.6163    &  \cellcolor[HTML]{FFEFEF}0.7080    &  \cellcolor[HTML]{FFEFEF}0.5586   &  \cellcolor[HTML]{FFEFEF}0.6004     &  \cellcolor[HTML]{FFEFEF}0.5834    &  \cellcolor[HTML]{FFEFEF}0.4265         &   \cellcolor[HTML]{FFEFEF}0.6883         &    \cellcolor[HTML]{FFEFEF}0.4926       &  \cellcolor[HTML]{FFEFEF}0.8228       &   \cellcolor[HTML]{FFEFEF}0.8422    \\
&   & \cellcolor[HTML]{F3FFF2}$E_{\xi}^{mean}\uparrow$ &  \cellcolor[HTML]{F3FFF2}0.7428     &    \cellcolor[HTML]{F3FFF2}0.7295  &  \cellcolor[HTML]{F3FFF2}0.8091    &  \cellcolor[HTML]{F3FFF2}0.6852   &     \cellcolor[HTML]{F3FFF2}0.7197  &    \cellcolor[HTML]{F3FFF2}0.7185  &        \cellcolor[HTML]{F3FFF2}0.5979   &      \cellcolor[HTML]{F3FFF2}0.8019      &      \cellcolor[HTML]{F3FFF2}0.6574     &         \cellcolor[HTML]{F3FFF2}0.9092      &    \cellcolor[HTML]{F3FFF2}0.9210                     \\\cdashline{2-14}[5pt/3pt]
& \multirow{4}{*}{MINET}     & \cellcolor[HTML]{DCF7FF}$MAE\downarrow$ &    \cellcolor[HTML]{DCF7FF}0.0726 &   \cellcolor[HTML]{DCF7FF}0.0680 &  \cellcolor[HTML]{DCF7FF}0.0627    &   \cellcolor[HTML]{DCF7FF}0.1113  &   \cellcolor[HTML]{DCF7FF}0.0764    &    \cellcolor[HTML]{DCF7FF}0.0742  &    \cellcolor[HTML]{DCF7FF}0.0871       &     \cellcolor[HTML]{DCF7FF}0.0681       &     \cellcolor[HTML]{DCF7FF}0.0752      &    \cellcolor[HTML]{DCF7FF}0.0349           &       \cellcolor[HTML]{DCF7FF}0.0324   \\
&      & \cellcolor[HTML]{FFFFE0}$S\uparrow$ & \cellcolor[HTML]{FFFFE0}0.7500      &  \cellcolor[HTML]{FFFFE0}0.7407    &  \cellcolor[HTML]{FFFFE0}0.7881    &   \cellcolor[HTML]{FFFFE0}0.7415  &  \cellcolor[HTML]{FFFFE0}0.7398     &  \cellcolor[HTML]{FFFFE0}0.7457    & \cellcolor[HTML]{FFFFE0}0.7095    &  \cellcolor[HTML]{FFFFE0}0.7693   &  \cellcolor[HTML]{FFFFE0}0.7572    &  \cellcolor[HTML]{FFFFE0}0.8721  &   \cellcolor[HTML]{FFFFE0}0.8772   \\
&      &  \cellcolor[HTML]{FFEFEF}$F_{\beta}^{mean}\uparrow$ & \cellcolor[HTML]{FFEFEF}0.6525      &  \cellcolor[HTML]{FFEFEF}0.6377    &  \cellcolor[HTML]{FFEFEF}0.7021    &  \cellcolor[HTML]{FFEFEF}0.6347   &  \cellcolor[HTML]{FFEFEF}0.6261     &  \cellcolor[HTML]{FFEFEF}0.6388    &  \cellcolor[HTML]{FFEFEF}0.5561         &   \cellcolor[HTML]{FFEFEF}0.6865         &    \cellcolor[HTML]{FFEFEF}0.6610       &  \cellcolor[HTML]{FFEFEF}0.8219       &   \cellcolor[HTML]{FFEFEF}0.8273    \\
&   & \cellcolor[HTML]{F3FFF2}$E_{\xi}^{mean}\uparrow$ &  \cellcolor[HTML]{F3FFF2}0.7764     &    \cellcolor[HTML]{F3FFF2}0.7535  &  \cellcolor[HTML]{F3FFF2}0.8095    &  \cellcolor[HTML]{F3FFF2}0.7676   &     \cellcolor[HTML]{F3FFF2}0.7827  &    \cellcolor[HTML]{F3FFF2}0.7586  &        \cellcolor[HTML]{F3FFF2}0.7134   &      \cellcolor[HTML]{F3FFF2}0.8011      &      \cellcolor[HTML]{F3FFF2}0.7999     &         \cellcolor[HTML]{F3FFF2}0.9110      &    \cellcolor[HTML]{F3FFF2}0.9140                     \\\cdashline{2-14}[5pt/3pt]
& \multirow{4}{*}{EDN}   & \cellcolor[HTML]{DCF7FF}$MAE\downarrow$ &    \cellcolor[HTML]{DCF7FF}0.0573 &   \cellcolor[HTML]{DCF7FF}0.0534 &  \cellcolor[HTML]{DCF7FF}0.0497    &   \cellcolor[HTML]{DCF7FF}0.0558  &   \cellcolor[HTML]{DCF7FF}0.0614    &    \cellcolor[HTML]{DCF7FF}0.0550  &    \cellcolor[HTML]{DCF7FF}0.0630       &     \cellcolor[HTML]{DCF7FF}0.0481       &     \cellcolor[HTML]{DCF7FF}0.0553      &    \cellcolor[HTML]{DCF7FF}0.0335           &       \cellcolor[HTML]{DCF7FF}0.0317   \\
&      & \cellcolor[HTML]{FFFFE0}$S\uparrow$ & \cellcolor[HTML]{FFFFE0}0.7719      &  \cellcolor[HTML]{FFFFE0}0.8021    &  \cellcolor[HTML]{FFFFE0}0.8213    &   \cellcolor[HTML]{FFFFE0}0.7814  &  \cellcolor[HTML]{FFFFE0}0.7606     &  \cellcolor[HTML]{FFFFE0}0.7960    & \cellcolor[HTML]{FFFFE0}0.7796    &  \cellcolor[HTML]{FFFFE0}0.8296   &  \cellcolor[HTML]{FFFFE0}0.8015    &  \cellcolor[HTML]{FFFFE0}0.8779  &   \cellcolor[HTML]{FFFFE0}0.8827   \\
&      &  \cellcolor[HTML]{FFEFEF}$F_{\beta}^{mean}\uparrow$ & \cellcolor[HTML]{FFEFEF}0.6829      &  \cellcolor[HTML]{FFEFEF}0.7220    &  \cellcolor[HTML]{FFEFEF}0.7541    &  \cellcolor[HTML]{FFEFEF}0.6864   &  \cellcolor[HTML]{FFEFEF}0.6491     &  \cellcolor[HTML]{FFEFEF}0.7095    &  \cellcolor[HTML]{FFEFEF}0.6745         &   \cellcolor[HTML]{FFEFEF}0.7618         &    \cellcolor[HTML]{FFEFEF}0.7136       &  \cellcolor[HTML]{FFEFEF}0.8280       &   \cellcolor[HTML]{FFEFEF}0.8337    \\
&   & \cellcolor[HTML]{F3FFF2}$E_{\xi}^{mean}\uparrow$ &  \cellcolor[HTML]{F3FFF2}0.7798     &    \cellcolor[HTML]{F3FFF2}0.8181  &  \cellcolor[HTML]{F3FFF2}0.8472    &  \cellcolor[HTML]{F3FFF2}0.7861   &     \cellcolor[HTML]{F3FFF2}0.7624  &    \cellcolor[HTML]{F3FFF2}0.8243  &        \cellcolor[HTML]{F3FFF2}0.7950   &      \cellcolor[HTML]{F3FFF2}0.8613      &      \cellcolor[HTML]{F3FFF2}0.8188     &         \cellcolor[HTML]{F3FFF2}0.9084      &    \cellcolor[HTML]{F3FFF2}0.9145                     \\\cdashline{2-14}[5pt/3pt]
& \multirow{4}{*}{AESINet}     & \cellcolor[HTML]{DCF7FF}$MAE\downarrow$ &    \cellcolor[HTML]{DCF7FF}0.0668 &   \cellcolor[HTML]{DCF7FF}0.0617 &  \cellcolor[HTML]{DCF7FF}0.0555    &   \cellcolor[HTML]{DCF7FF}0.0702  &   \cellcolor[HTML]{DCF7FF}0.0616    &    \cellcolor[HTML]{DCF7FF}0.0663  &    \cellcolor[HTML]{DCF7FF}0.0811       &     \cellcolor[HTML]{DCF7FF}0.0606       &     \cellcolor[HTML]{DCF7FF}0.0904      &    \cellcolor[HTML]{DCF7FF}0.0328           &       \cellcolor[HTML]{DCF7FF}0.0314   \\
&      & \cellcolor[HTML]{FFFFE0}$S\uparrow$ & \cellcolor[HTML]{FFFFE0}0.7268      &  \cellcolor[HTML]{FFFFE0}0.7388    &  \cellcolor[HTML]{FFFFE0}0.7920    &   \cellcolor[HTML]{FFFFE0}0.7352  &  \cellcolor[HTML]{FFFFE0}0.7430     &  \cellcolor[HTML]{FFFFE0}0.7140    & \cellcolor[HTML]{FFFFE0}0.6744    &  \cellcolor[HTML]{FFFFE0}0.7607   &  \cellcolor[HTML]{FFFFE0}0.7027    &  \cellcolor[HTML]{FFFFE0}0.8691  &   \cellcolor[HTML]{FFFFE0}0.8714   \\
&      &  \cellcolor[HTML]{FFEFEF}$F_{\beta}^{mean}\uparrow$ & \cellcolor[HTML]{FFEFEF}0.7325      &  \cellcolor[HTML]{FFEFEF}0.6284    &  \cellcolor[HTML]{FFEFEF}0.7065    &  \cellcolor[HTML]{FFEFEF}0.6322   &  \cellcolor[HTML]{FFEFEF}0.6343     &  \cellcolor[HTML]{FFEFEF}0.5796    &  \cellcolor[HTML]{FFEFEF}0.4997         &   \cellcolor[HTML]{FFEFEF}0.6690         &    \cellcolor[HTML]{FFEFEF}0.5768       &  \cellcolor[HTML]{FFEFEF}0.8292       &   \cellcolor[HTML]{FFEFEF}0.8306    \\
&   & \cellcolor[HTML]{F3FFF2}$E_{\xi}^{mean}\uparrow$ &  \cellcolor[HTML]{F3FFF2}0.7325     &    \cellcolor[HTML]{F3FFF2}0.7438  &  \cellcolor[HTML]{F3FFF2}0.8298    &  \cellcolor[HTML]{F3FFF2}0.7687   &     \cellcolor[HTML]{F3FFF2}0.7827  &    \cellcolor[HTML]{F3FFF2}0.7054  &        \cellcolor[HTML]{F3FFF2}0.6570   &      \cellcolor[HTML]{F3FFF2}0.7842      &      \cellcolor[HTML]{F3FFF2}0.7159     &         \cellcolor[HTML]{F3FFF2}0.9140      &    \cellcolor[HTML]{F3FFF2}0.9160                     \\\cdashline{2-14}[5pt/3pt]
& \multirow{4}{*}{BAFSNet}      & \cellcolor[HTML]{DCF7FF}$MAE\downarrow$ &    \cellcolor[HTML]{DCF7FF}0.0620 &   \cellcolor[HTML]{DCF7FF}0.0583 &  \cellcolor[HTML]{DCF7FF}0.0501    &   \cellcolor[HTML]{DCF7FF}0.0732  &   \cellcolor[HTML]{DCF7FF}0.0611    &    \cellcolor[HTML]{DCF7FF}0.0669  &    \cellcolor[HTML]{DCF7FF}0.0711       &     \cellcolor[HTML]{DCF7FF}0.0511       &     \cellcolor[HTML]{DCF7FF}0.0770      &    \cellcolor[HTML]{DCF7FF}0.0315           &       \cellcolor[HTML]{DCF7FF}0.0298   \\
&      & \cellcolor[HTML]{FFFFE0}$S\uparrow$ & \cellcolor[HTML]{FFFFE0}0.7444      &  \cellcolor[HTML]{FFFFE0}0.7692    &  \cellcolor[HTML]{FFFFE0}0.8183    &   \cellcolor[HTML]{FFFFE0}0.7321  &  \cellcolor[HTML]{FFFFE0}0.7620     &  \cellcolor[HTML]{FFFFE0}0.7444    & \cellcolor[HTML]{FFFFE0}0.7206    &  \cellcolor[HTML]{FFFFE0}0.8151   &  \cellcolor[HTML]{FFFFE0}0.6784    &  \cellcolor[HTML]{FFFFE0}0.8825  &   \cellcolor[HTML]{FFFFE0}0.8857   \\
&      &  \cellcolor[HTML]{FFEFEF}$F_{\beta}^{mean}\uparrow$ & \cellcolor[HTML]{FFEFEF}0.6448      &  \cellcolor[HTML]{FFEFEF}0.6791    &  \cellcolor[HTML]{FFEFEF}0.7588    &  \cellcolor[HTML]{FFEFEF}0.6132   &  \cellcolor[HTML]{FFEFEF}0.6577     &  \cellcolor[HTML]{FFEFEF}0.6305    &  \cellcolor[HTML]{FFEFEF}0.5835         &   \cellcolor[HTML]{FFEFEF}0.7468         &    \cellcolor[HTML]{FFEFEF}0.5322       &  \cellcolor[HTML]{FFEFEF}0.8404       &   \cellcolor[HTML]{FFEFEF}0.8422    \\
&   & \cellcolor[HTML]{F3FFF2}$E_{\xi}^{mean}\uparrow$ &  \cellcolor[HTML]{F3FFF2}0.7428     &    \cellcolor[HTML]{F3FFF2}0.7786  &  \cellcolor[HTML]{F3FFF2}0.8511    &  \cellcolor[HTML]{F3FFF2}0.7413   &     \cellcolor[HTML]{F3FFF2}0.7814  &    \cellcolor[HTML]{F3FFF2}0.7510  &        \cellcolor[HTML]{F3FFF2}0.7341   &      \cellcolor[HTML]{F3FFF2}0.8485      &      \cellcolor[HTML]{F3FFF2}0.6644     &         \cellcolor[HTML]{F3FFF2}0.9199      &    \cellcolor[HTML]{F3FFF2}0.9273                     \\\cdashline{2-14}[5pt/3pt]
& \multirow{4}{*}{DC-Net}     & \cellcolor[HTML]{DCF7FF}$MAE\downarrow$ &    \cellcolor[HTML]{DCF7FF}0.0753 &   \cellcolor[HTML]{DCF7FF}0.0698 &  \cellcolor[HTML]{DCF7FF}0.0747    &   \cellcolor[HTML]{DCF7FF}0.0732  &   \cellcolor[HTML]{DCF7FF}0.0779    &    \cellcolor[HTML]{DCF7FF}0.0773  &    \cellcolor[HTML]{DCF7FF}0.0808       &     \cellcolor[HTML]{DCF7FF}0.0573       &     \cellcolor[HTML]{DCF7FF}0.0749      &    \cellcolor[HTML]{DCF7FF}0.0321           &       \cellcolor[HTML]{DCF7FF}0.0315   \\
&      & \cellcolor[HTML]{FFFFE0}$S\uparrow$ & \cellcolor[HTML]{FFFFE0}0.7010      &  \cellcolor[HTML]{FFFFE0}0.7565    &  \cellcolor[HTML]{FFFFE0}0.7624    &   \cellcolor[HTML]{FFFFE0}0.7311  &  \cellcolor[HTML]{FFFFE0}0.7168     &  \cellcolor[HTML]{FFFFE0}0.7263    & \cellcolor[HTML]{FFFFE0}0.7325    &  \cellcolor[HTML]{FFFFE0}0.7980   &  \cellcolor[HTML]{FFFFE0}0.7577    &  \cellcolor[HTML]{FFFFE0}0.8827  &   \cellcolor[HTML]{FFFFE0}0.8817   \\
&      &  \cellcolor[HTML]{FFEFEF}$F_{\beta}^{mean}\uparrow$ & \cellcolor[HTML]{FFEFEF}0.5550      &  \cellcolor[HTML]{FFEFEF}0.6553    &  \cellcolor[HTML]{FFEFEF}0.6694    &  \cellcolor[HTML]{FFEFEF}0.6116   &  \cellcolor[HTML]{FFEFEF}0.5838     &  \cellcolor[HTML]{FFEFEF}0.6050    &  \cellcolor[HTML]{FFEFEF}0.6049         &   \cellcolor[HTML]{FFEFEF}0.7197         &    \cellcolor[HTML]{FFEFEF}0.6603       &  \cellcolor[HTML]{FFEFEF}0.8399       &   \cellcolor[HTML]{FFEFEF}0.8382    \\
&   & \cellcolor[HTML]{F3FFF2}$E_{\xi}^{mean}\uparrow$ &  \cellcolor[HTML]{F3FFF2}0.6834     &    \cellcolor[HTML]{F3FFF2}0.7663  &  \cellcolor[HTML]{F3FFF2}0.7878    &  \cellcolor[HTML]{F3FFF2}0.7386   &     \cellcolor[HTML]{F3FFF2}0.7144  &    \cellcolor[HTML]{F3FFF2}0.7466  &        \cellcolor[HTML]{F3FFF2}0.7404   &      \cellcolor[HTML]{F3FFF2}0.8300      &      \cellcolor[HTML]{F3FFF2}0.7746     &         \cellcolor[HTML]{F3FFF2}0.9207       &    \cellcolor[HTML]{F3FFF2}0.9153                    \\\cdashline{2-14}[5pt/3pt]
& \multirow{4}{*}{RNet}      & \cellcolor[HTML]{DCF7FF}$MAE\downarrow$ &    \cellcolor[HTML]{DCF7FF}0.0614 &   \cellcolor[HTML]{DCF7FF}0.0632 &  \cellcolor[HTML]{DCF7FF}0.0574    &   \cellcolor[HTML]{DCF7FF}0.0714  &   \cellcolor[HTML]{DCF7FF}0.0687    &    \cellcolor[HTML]{DCF7FF}0.0673  &    \cellcolor[HTML]{DCF7FF}0.0815       &     \cellcolor[HTML]{DCF7FF}0.0606       &     \cellcolor[HTML]{DCF7FF}0.0893      &    \cellcolor[HTML]{DCF7FF}0.0372           &       \cellcolor[HTML]{DCF7FF}0.0329   \\
&      & \cellcolor[HTML]{FFFFE0}$S\uparrow$ & \cellcolor[HTML]{FFFFE0}0.7471      &  \cellcolor[HTML]{FFFFE0}0.7544    &  \cellcolor[HTML]{FFFFE0}0.7898    &   \cellcolor[HTML]{FFFFE0}0.7371  &  \cellcolor[HTML]{FFFFE0}0.7117     &  \cellcolor[HTML]{FFFFE0}0.7435    & \cellcolor[HTML]{FFFFE0}0.6766    &  \cellcolor[HTML]{FFFFE0}0.7768   &  \cellcolor[HTML]{FFFFE0}0.6738    &  \cellcolor[HTML]{FFFFE0}0.8668  &   \cellcolor[HTML]{FFFFE0}0.8762   \\
&      &  \cellcolor[HTML]{FFEFEF}$F_{\beta}^{mean}\uparrow$ & \cellcolor[HTML]{FFEFEF}0.6412      &  \cellcolor[HTML]{FFEFEF}0.6338    &  \cellcolor[HTML]{FFEFEF}0.7075    &  \cellcolor[HTML]{FFEFEF}0.6094   &  \cellcolor[HTML]{FFEFEF}0.5703     &  \cellcolor[HTML]{FFEFEF}0.6109    &  \cellcolor[HTML]{FFEFEF}0.4930         &   \cellcolor[HTML]{FFEFEF}0.6848         &    \cellcolor[HTML]{FFEFEF}0.5208       &  \cellcolor[HTML]{FFEFEF}0.8221       &   \cellcolor[HTML]{FFEFEF}0.8313    \\
&   & \cellcolor[HTML]{F3FFF2}$E_{\xi}^{mean}\uparrow$ &  \cellcolor[HTML]{F3FFF2}0.7567     &    \cellcolor[HTML]{F3FFF2}0.7575  &  \cellcolor[HTML]{F3FFF2}0.8011    &  \cellcolor[HTML]{F3FFF2}0.7297   &     \cellcolor[HTML]{F3FFF2}0.6922  &    \cellcolor[HTML]{F3FFF2}0.7408  &        \cellcolor[HTML]{F3FFF2}0.6386   &      \cellcolor[HTML]{F3FFF2}0.7991      &      \cellcolor[HTML]{F3FFF2}0.6689     &         \cellcolor[HTML]{F3FFF2}0.9068      &    \cellcolor[HTML]{F3FFF2}0.9144                     \\\cdashline{2-14}[5pt/3pt]
& \multirow{4}{*}{Average}     & \cellcolor[HTML]{DCF7FF}$MAE\downarrow$ &    \cellcolor[HTML]{DCF7FF}0.0665 &   \cellcolor[HTML]{DCF7FF}0.0631 &  \cellcolor[HTML]{DCF7FF}0.0582    &   \cellcolor[HTML]{DCF7FF}0.0764  &   \cellcolor[HTML]{DCF7FF}0.0681    &    \cellcolor[HTML]{DCF7FF}0.0698  &    \cellcolor[HTML]{DCF7FF}0.0790       &     \cellcolor[HTML]{DCF7FF}0.0581       &     \cellcolor[HTML]{DCF7FF}0.0832      &    \cellcolor[HTML]{DCF7FF}0.0339           &       \cellcolor[HTML]{DCF7FF}0.0315   \\
&      & \cellcolor[HTML]{FFFFE0}$S\uparrow$ & \cellcolor[HTML]{FFFFE0}0.7401      &  \cellcolor[HTML]{FFFFE0}0.7578    &  \cellcolor[HTML]{FFFFE0}0.7948    &   \cellcolor[HTML]{FFFFE0}0.7387  &  \cellcolor[HTML]{FFFFE0}0.7382     &  \cellcolor[HTML]{FFFFE0}0.7415    & \cellcolor[HTML]{FFFFE0}0.7048    &  \cellcolor[HTML]{FFFFE0}0.7905   &  \cellcolor[HTML]{FFFFE0}0.7192    &  \cellcolor[HTML]{FFFFE0}0.8749  &   \cellcolor[HTML]{FFFFE0}0.8796   \\
&      &  \cellcolor[HTML]{FFEFEF}$F_{\beta}^{mean}\uparrow$ & \cellcolor[HTML]{FFEFEF}0.6309      &  \cellcolor[HTML]{FFEFEF}0.6532    &  \cellcolor[HTML]{FFEFEF}0.7152    &  \cellcolor[HTML]{FFEFEF}0.6209   &  \cellcolor[HTML]{FFEFEF}0.6174     &  \cellcolor[HTML]{FFEFEF}0.6225    &  \cellcolor[HTML]{FFEFEF}0.5483         &   \cellcolor[HTML]{FFEFEF}0.7081         &    \cellcolor[HTML]{FFEFEF}0.5939       &  \cellcolor[HTML]{FFEFEF}0.8292       &   \cellcolor[HTML]{FFEFEF}0.8351    \\
&   & \cellcolor[HTML]{F3FFF2}$E_{\xi}^{mean}\uparrow$ &  \cellcolor[HTML]{F3FFF2}0.7449     &    \cellcolor[HTML]{F3FFF2}0.7639  &  \cellcolor[HTML]{F3FFF2}0.8194    &  \cellcolor[HTML]{F3FFF2}0.7453   &     \cellcolor[HTML]{F3FFF2}0.7479  &    \cellcolor[HTML]{F3FFF2}0.7493  &        \cellcolor[HTML]{F3FFF2}0.6966   &      \cellcolor[HTML]{F3FFF2}0.8180      &      \cellcolor[HTML]{F3FFF2}0.7286     &         \cellcolor[HTML]{F3FFF2}0.9129      &    \cellcolor[HTML]{F3FFF2}0.9175                     \\\hline\hline
\multirow{32}{*}{Real} &  \multirow{4}{*}{F3Net}      & \cellcolor[HTML]{DCF7FF}$MAE\downarrow$ &    \cellcolor[HTML]{DCF7FF}0.0346 &   \cellcolor[HTML]{DCF7FF}0.0490 &  \cellcolor[HTML]{DCF7FF}0.0372    &   \cellcolor[HTML]{DCF7FF}0.1494  &   \cellcolor[HTML]{DCF7FF}0.0351    &    \cellcolor[HTML]{DCF7FF}0.0469  &    \cellcolor[HTML]{DCF7FF}0.0827       &     \cellcolor[HTML]{DCF7FF}0.0476       &     \cellcolor[HTML]{DCF7FF}0.1493      &    \cellcolor[HTML]{DCF7FF}0.0248           &       \cellcolor[HTML]{DCF7FF}0.0196   \\
&      & \cellcolor[HTML]{FFFFE0}$S\uparrow$ & \cellcolor[HTML]{FFFFE0}0.8481      &  \cellcolor[HTML]{FFFFE0}0.8006    &  \cellcolor[HTML]{FFFFE0}0.8332    &   \cellcolor[HTML]{FFFFE0}0.7559  &  \cellcolor[HTML]{FFFFE0}0.8468     &  \cellcolor[HTML]{FFFFE0}0.8305    & \cellcolor[HTML]{FFFFE0}0.6354    &  \cellcolor[HTML]{FFFFE0}0.7963   &  \cellcolor[HTML]{FFFFE0}0.5801    &  \cellcolor[HTML]{FFFFE0}0.8982  &   \cellcolor[HTML]{FFFFE0}0.9133   \\
&      &  \cellcolor[HTML]{FFEFEF}$F_{\beta}^{mean}\uparrow$ & \cellcolor[HTML]{FFEFEF}0.7630      &  \cellcolor[HTML]{FFEFEF}0.6853    &  \cellcolor[HTML]{FFEFEF}0.7384    &  \cellcolor[HTML]{FFEFEF}0.6776   &  \cellcolor[HTML]{FFEFEF}0.7663     &  \cellcolor[HTML]{FFEFEF}0.7384    &  \cellcolor[HTML]{FFEFEF}0.3958         &   \cellcolor[HTML]{FFEFEF}0.6796         &    \cellcolor[HTML]{FFEFEF}0.3416       &  \cellcolor[HTML]{FFEFEF}0.8375       &   \cellcolor[HTML]{FFEFEF}0.8611    \\
&   & \cellcolor[HTML]{F3FFF2}$E_{\xi}^{mean}\uparrow$ &  \cellcolor[HTML]{F3FFF2}0.8563     &    \cellcolor[HTML]{F3FFF2}0.7876  &  \cellcolor[HTML]{F3FFF2}0.8386    &  \cellcolor[HTML]{F3FFF2}0.7722   &     \cellcolor[HTML]{F3FFF2}0.8606  &    \cellcolor[HTML]{F3FFF2}0.8462  &        \cellcolor[HTML]{F3FFF2}0.5663   &      \cellcolor[HTML]{F3FFF2}0.7870      &      \cellcolor[HTML]{F3FFF2}0.5444     &         \cellcolor[HTML]{F3FFF2}0.9198      &    \cellcolor[HTML]{F3FFF2}0.9383                     \\\cdashline{2-14}[5pt/3pt]
& \multirow{4}{*}{MINET}     & \cellcolor[HTML]{DCF7FF}$MAE\downarrow$ &    \cellcolor[HTML]{DCF7FF}0.0373 &   \cellcolor[HTML]{DCF7FF}0.0496 &  \cellcolor[HTML]{DCF7FF}0.0366    &   \cellcolor[HTML]{DCF7FF}0.0601  &   \cellcolor[HTML]{DCF7FF}0.0449    &    \cellcolor[HTML]{DCF7FF}0.0355  &    \cellcolor[HTML]{DCF7FF}0.0903       &     \cellcolor[HTML]{DCF7FF}0.0514       &     \cellcolor[HTML]{DCF7FF}0.1130      &    \cellcolor[HTML]{DCF7FF}0.0232           &       \cellcolor[HTML]{DCF7FF}0.0241   \\
&      & \cellcolor[HTML]{FFFFE0}$S\uparrow$ & \cellcolor[HTML]{FFFFE0}0.8652      &  \cellcolor[HTML]{FFFFE0}0.8003    &  \cellcolor[HTML]{FFFFE0}0.8527    &   \cellcolor[HTML]{FFFFE0}0.8330  &  \cellcolor[HTML]{FFFFE0}0.8451     &  \cellcolor[HTML]{FFFFE0}0.8559    & \cellcolor[HTML]{FFFFE0}0.7073    &  \cellcolor[HTML]{FFFFE0}0.8046   &  \cellcolor[HTML]{FFFFE0}0.7316    &  \cellcolor[HTML]{FFFFE0}0.9007  &   \cellcolor[HTML]{FFFFE0}0.9038   \\
&      &  \cellcolor[HTML]{FFEFEF}$F_{\beta}^{mean}\uparrow$ & \cellcolor[HTML]{FFEFEF}0.7942      &  \cellcolor[HTML]{FFEFEF}0.7076    &  \cellcolor[HTML]{FFEFEF}0.7823    &  \cellcolor[HTML]{FFEFEF}0.7394   &  \cellcolor[HTML]{FFEFEF}0.7686     &  \cellcolor[HTML]{FFEFEF}0.7721    &  \cellcolor[HTML]{FFEFEF}0.5218         &   \cellcolor[HTML]{FFEFEF}0.7010         &    \cellcolor[HTML]{FFEFEF}0.6055       &  \cellcolor[HTML]{FFEFEF}0.8470       &   \cellcolor[HTML]{FFEFEF}0.8445    \\
&   & \cellcolor[HTML]{F3FFF2}$E_{\xi}^{mean}\uparrow$ &  \cellcolor[HTML]{F3FFF2}0.8928     &    \cellcolor[HTML]{F3FFF2}0.8063  &  \cellcolor[HTML]{F3FFF2}0.8683    &  \cellcolor[HTML]{F3FFF2}0.8529   &     \cellcolor[HTML]{F3FFF2}0.8787  &    \cellcolor[HTML]{F3FFF2}0.8849  &        \cellcolor[HTML]{F3FFF2}0.7116   &      \cellcolor[HTML]{F3FFF2}0.8087      &      \cellcolor[HTML]{F3FFF2}0.7545     &         \cellcolor[HTML]{F3FFF2}0.9282      &    \cellcolor[HTML]{F3FFF2}0.9250                     \\\cdashline{2-14}[5pt/3pt]
& \multirow{4}{*}{EDN}   & \cellcolor[HTML]{DCF7FF}$MAE\downarrow$ &    \cellcolor[HTML]{DCF7FF}0.0295 &   \cellcolor[HTML]{DCF7FF}0.0301 &  \cellcolor[HTML]{DCF7FF}0.0240    &   \cellcolor[HTML]{DCF7FF}0.0261  &   \cellcolor[HTML]{DCF7FF}0.0278    &    \cellcolor[HTML]{DCF7FF}0.0312  &    \cellcolor[HTML]{DCF7FF}0.0422       &     \cellcolor[HTML]{DCF7FF}0.0337       &     \cellcolor[HTML]{DCF7FF}0.0454      &    \cellcolor[HTML]{DCF7FF}0.0223           &       \cellcolor[HTML]{DCF7FF}0.0183   \\
&      & \cellcolor[HTML]{FFFFE0}$S\uparrow$ & \cellcolor[HTML]{FFFFE0}0.8857      &  \cellcolor[HTML]{FFFFE0}0.8887    &  \cellcolor[HTML]{FFFFE0}0.8921    &   \cellcolor[HTML]{FFFFE0}0.8931  &  \cellcolor[HTML]{FFFFE0}0.8653     &  \cellcolor[HTML]{FFFFE0}0.8878    & \cellcolor[HTML]{FFFFE0}0.8244    &  \cellcolor[HTML]{FFFFE0}0.8717   &  \cellcolor[HTML]{FFFFE0}0.8377    &  \cellcolor[HTML]{FFFFE0}0.9097  &   \cellcolor[HTML]{FFFFE0}0.9173   \\
&      &  \cellcolor[HTML]{FFEFEF}$F_{\beta}^{mean}\uparrow$ & \cellcolor[HTML]{FFEFEF}0.8170      &  \cellcolor[HTML]{FFEFEF}0.8178    &  \cellcolor[HTML]{FFEFEF}0.8197    &  \cellcolor[HTML]{FFEFEF}0.8200   &  \cellcolor[HTML]{FFEFEF}0.7820     &  \cellcolor[HTML]{FFEFEF}0.8188    &  \cellcolor[HTML]{FFEFEF}0.7097         &   \cellcolor[HTML]{FFEFEF}0.7942         &    \cellcolor[HTML]{FFEFEF}0.7432       &  \cellcolor[HTML]{FFEFEF}0.8520       &   \cellcolor[HTML]{FFEFEF}0.8634    \\
&   & \cellcolor[HTML]{F3FFF2}$E_{\xi}^{mean}\uparrow$ &  \cellcolor[HTML]{F3FFF2}0.8993     &    \cellcolor[HTML]{F3FFF2}0.9083  &  \cellcolor[HTML]{F3FFF2}0.9111    &  \cellcolor[HTML]{F3FFF2}0.9088   &     \cellcolor[HTML]{F3FFF2}0.8818  &    \cellcolor[HTML]{F3FFF2}0.9120  &        \cellcolor[HTML]{F3FFF2}0.8261   &      \cellcolor[HTML]{F3FFF2}0.8875      &      \cellcolor[HTML]{F3FFF2}0.8514     &         \cellcolor[HTML]{F3FFF2}0.9272      &    \cellcolor[HTML]{F3FFF2}0.9395                     \\\cdashline{2-14}[5pt/3pt]
& \multirow{4}{*}{AESINet}     & \cellcolor[HTML]{DCF7FF}$MAE\downarrow$ &    \cellcolor[HTML]{DCF7FF}0.0368 &   \cellcolor[HTML]{DCF7FF}0.0387 &  \cellcolor[HTML]{DCF7FF}0.0389    &   \cellcolor[HTML]{DCF7FF}0.0785  &   \cellcolor[HTML]{DCF7FF}0.0323    &    \cellcolor[HTML]{DCF7FF}0.0393  &    \cellcolor[HTML]{DCF7FF}0.1011       &     \cellcolor[HTML]{DCF7FF}0.0478       &     \cellcolor[HTML]{DCF7FF}0.2105      &    \cellcolor[HTML]{DCF7FF}0.0253           &       \cellcolor[HTML]{DCF7FF}0.0179   \\
&      & \cellcolor[HTML]{FFFFE0}$S\uparrow$ & \cellcolor[HTML]{FFFFE0}0.8459      &  \cellcolor[HTML]{FFFFE0}0.8071    &  \cellcolor[HTML]{FFFFE0}0.8556    &   \cellcolor[HTML]{FFFFE0}0.8004  &  \cellcolor[HTML]{FFFFE0}0.8616     &  \cellcolor[HTML]{FFFFE0}0.8213    & \cellcolor[HTML]{FFFFE0}0.6912    &  \cellcolor[HTML]{FFFFE0}0.7921   &  \cellcolor[HTML]{FFFFE0}0.6237    &  \cellcolor[HTML]{FFFFE0}0.8916  &   \cellcolor[HTML]{FFFFE0}0.9097   \\
&      &  \cellcolor[HTML]{FFEFEF}$F_{\beta}^{mean}\uparrow$ & \cellcolor[HTML]{FFEFEF}0.7750      &  \cellcolor[HTML]{FFEFEF}0.7095    &  \cellcolor[HTML]{FFEFEF}0.7710    &  \cellcolor[HTML]{FFEFEF}0.7049   &  \cellcolor[HTML]{FFEFEF}0.8015     &  \cellcolor[HTML]{FFEFEF}0.7352    &  \cellcolor[HTML]{FFEFEF}0.5422         &   \cellcolor[HTML]{FFEFEF}0.6900         &    \cellcolor[HTML]{FFEFEF}0.5163       &  \cellcolor[HTML]{FFEFEF}0.8441       &   \cellcolor[HTML]{FFEFEF}0.8694    \\
&   & \cellcolor[HTML]{F3FFF2}$E_{\xi}^{mean}\uparrow$ &  \cellcolor[HTML]{F3FFF2}0.8689     &    \cellcolor[HTML]{F3FFF2}0.8160  &  \cellcolor[HTML]{F3FFF2}0.8841    &  \cellcolor[HTML]{F3FFF2}0.8271   &     \cellcolor[HTML]{F3FFF2}0.8953  &    \cellcolor[HTML]{F3FFF2}0.8429  &        \cellcolor[HTML]{F3FFF2}0.6977   &      \cellcolor[HTML]{F3FFF2}0.8042      &      \cellcolor[HTML]{F3FFF2}0.6293     &         \cellcolor[HTML]{F3FFF2}0.9297      &    \cellcolor[HTML]{F3FFF2}0.9441                     \\\cdashline{2-14}[5pt/3pt]
& \multirow{4}{*}{BAFSNet}      & \cellcolor[HTML]{DCF7FF}$MAE\downarrow$ &    \cellcolor[HTML]{DCF7FF}0.0300 &   \cellcolor[HTML]{DCF7FF}0.0355 &  \cellcolor[HTML]{DCF7FF}0.0278    &   \cellcolor[HTML]{DCF7FF}0.1199  &   \cellcolor[HTML]{DCF7FF}0.0339    &    \cellcolor[HTML]{DCF7FF}0.0354  &    \cellcolor[HTML]{DCF7FF}0.0649       &     \cellcolor[HTML]{DCF7FF}0.0339       &     \cellcolor[HTML]{DCF7FF}0.0765      &    \cellcolor[HTML]{DCF7FF}0.0198           &       \cellcolor[HTML]{DCF7FF}0.0197   \\
&      & \cellcolor[HTML]{FFFFE0}$S\uparrow$ & \cellcolor[HTML]{FFFFE0}0.8726      &  \cellcolor[HTML]{FFFFE0}0.8339    &  \cellcolor[HTML]{FFFFE0}0.8870    &   \cellcolor[HTML]{FFFFE0}0.7609  &  \cellcolor[HTML]{FFFFE0}0.8631     &  \cellcolor[HTML]{FFFFE0}0.8685    & \cellcolor[HTML]{FFFFE0}0.7154    &  \cellcolor[HTML]{FFFFE0}0.8546   &  \cellcolor[HTML]{FFFFE0}0.6545    &  \cellcolor[HTML]{FFFFE0}0.9088  &   \cellcolor[HTML]{FFFFE0}0.9132   \\
&      &  \cellcolor[HTML]{FFEFEF}$F_{\beta}^{mean}\uparrow$ & \cellcolor[HTML]{FFEFEF}0.8139      &  \cellcolor[HTML]{FFEFEF}0.7575    &  \cellcolor[HTML]{FFEFEF}0.8339    &  \cellcolor[HTML]{FFEFEF}0.6559   &  \cellcolor[HTML]{FFEFEF}0.7910     &  \cellcolor[HTML]{FFEFEF}0.7924    &  \cellcolor[HTML]{FFEFEF}0.5490         &   \cellcolor[HTML]{FFEFEF}0.7776         &    \cellcolor[HTML]{FFEFEF}0.4459       &  \cellcolor[HTML]{FFEFEF}0.8565       &   \cellcolor[HTML]{FFEFEF}0.8620    \\
&   & \cellcolor[HTML]{F3FFF2}$E_{\xi}^{mean}\uparrow$ &  \cellcolor[HTML]{F3FFF2}0.8978     &    \cellcolor[HTML]{F3FFF2}0.8530  &  \cellcolor[HTML]{F3FFF2}0.9206    &  \cellcolor[HTML]{F3FFF2}0.7805   &     \cellcolor[HTML]{F3FFF2}0.8899  &    \cellcolor[HTML]{F3FFF2}0.8939  &        \cellcolor[HTML]{F3FFF2}0.6971   &      \cellcolor[HTML]{F3FFF2}0.8786      &      \cellcolor[HTML]{F3FFF2}0.6062     &         \cellcolor[HTML]{F3FFF2}0.9307      &    \cellcolor[HTML]{F3FFF2}0.9395                     \\\cdashline{2-14}[5pt/3pt]
& \multirow{4}{*}{DC-Net}     & \cellcolor[HTML]{DCF7FF}$MAE\downarrow$ &    \cellcolor[HTML]{DCF7FF}0.0399 &   \cellcolor[HTML]{DCF7FF}0.0451 &  \cellcolor[HTML]{DCF7FF}0.0538    &   \cellcolor[HTML]{DCF7FF}0.0458  &   \cellcolor[HTML]{DCF7FF}0.0440    &    \cellcolor[HTML]{DCF7FF}0.0411  &    \cellcolor[HTML]{DCF7FF}0.1174       &     \cellcolor[HTML]{DCF7FF}0.0483       &     \cellcolor[HTML]{DCF7FF}0.0693      &    \cellcolor[HTML]{DCF7FF}0.0261           &       \cellcolor[HTML]{DCF7FF}0.0194   \\
&      & \cellcolor[HTML]{FFFFE0}$S\uparrow$ & \cellcolor[HTML]{FFFFE0}0.8481      &  \cellcolor[HTML]{FFFFE0}0.8214    &  \cellcolor[HTML]{FFFFE0}0.8084    &   \cellcolor[HTML]{FFFFE0}0.8503  &  \cellcolor[HTML]{FFFFE0}0.8307     &  \cellcolor[HTML]{FFFFE0}0.8372    & \cellcolor[HTML]{FFFFE0}0.7422    &  \cellcolor[HTML]{FFFFE0}0.8242   &  \cellcolor[HTML]{FFFFE0}0.7786    &  \cellcolor[HTML]{FFFFE0}0.8960  &   \cellcolor[HTML]{FFFFE0}0.9079   \\
&      &  \cellcolor[HTML]{FFEFEF}$F_{\beta}^{mean}\uparrow$ & \cellcolor[HTML]{FFEFEF}0.7675      &  \cellcolor[HTML]{FFEFEF}0.7225    &  \cellcolor[HTML]{FFEFEF}0.7169    &  \cellcolor[HTML]{FFEFEF}0.7747   &  \cellcolor[HTML]{FFEFEF}0.7380     &  \cellcolor[HTML]{FFEFEF}0.7524    &  \cellcolor[HTML]{FFEFEF}0.6228         &   \cellcolor[HTML]{FFEFEF}0.7237         &    \cellcolor[HTML]{FFEFEF}0.6724       &  \cellcolor[HTML]{FFEFEF}0.8444       &   \cellcolor[HTML]{FFEFEF}0.8611    \\
&   & \cellcolor[HTML]{F3FFF2}$E_{\xi}^{mean}\uparrow$ &  \cellcolor[HTML]{F3FFF2}0.8686     &    \cellcolor[HTML]{F3FFF2}0.8297  &  \cellcolor[HTML]{F3FFF2}0.8251    &  \cellcolor[HTML]{F3FFF2}0.8635   &     \cellcolor[HTML]{F3FFF2}0.8505  &    \cellcolor[HTML]{F3FFF2}0.8550  &        \cellcolor[HTML]{F3FFF2}0.7455   &      \cellcolor[HTML]{F3FFF2}0.8277      &      \cellcolor[HTML]{F3FFF2}0.7904     &         \cellcolor[HTML]{F3FFF2}0.9249      &    \cellcolor[HTML]{F3FFF2}0.9356                     \\\cdashline{2-14}[5pt/3pt]
& \multirow{4}{*}{R-Net}    & \cellcolor[HTML]{DCF7FF}$MAE\downarrow$ &    \cellcolor[HTML]{DCF7FF}0.0380 &   \cellcolor[HTML]{DCF7FF}0.0450 &  \cellcolor[HTML]{DCF7FF}0.0305    &   \cellcolor[HTML]{DCF7FF}0.0545  &   \cellcolor[HTML]{DCF7FF}0.0385    &    \cellcolor[HTML]{DCF7FF}0.0458  &    \cellcolor[HTML]{DCF7FF}0.0719       &     \cellcolor[HTML]{DCF7FF}0.0452       &     \cellcolor[HTML]{DCF7FF}0.1044      &    \cellcolor[HTML]{DCF7FF}0.0260           &       \cellcolor[HTML]{DCF7FF}0.0189   \\
&      & \cellcolor[HTML]{FFFFE0}$S\uparrow$ & \cellcolor[HTML]{FFFFE0}0.8581      &  \cellcolor[HTML]{FFFFE0}0.8103    &  \cellcolor[HTML]{FFFFE0}0.8763    &   \cellcolor[HTML]{FFFFE0}0.8225  &  \cellcolor[HTML]{FFFFE0}0.7968     &  \cellcolor[HTML]{FFFFE0}0.8629    & \cellcolor[HTML]{FFFFE0}0.7116    &  \cellcolor[HTML]{FFFFE0}0.8014   &  \cellcolor[HTML]{FFFFE0}0.6537    &  \cellcolor[HTML]{FFFFE0}0.8938  &   \cellcolor[HTML]{FFFFE0}0.9150   \\
&      &  \cellcolor[HTML]{FFEFEF}$F_{\beta}^{mean}\uparrow$ & \cellcolor[HTML]{FFEFEF}0.7827      &  \cellcolor[HTML]{FFEFEF}0.6971    &  \cellcolor[HTML]{FFEFEF}0.8165    &  \cellcolor[HTML]{FFEFEF}0.7191   &  \cellcolor[HTML]{FFEFEF}0.6750     &  \cellcolor[HTML]{FFEFEF}0.7866    &  \cellcolor[HTML]{FFEFEF}0.5377         &   \cellcolor[HTML]{FFEFEF}0.6830         &    \cellcolor[HTML]{FFEFEF}0.4582       &  \cellcolor[HTML]{FFEFEF}0.8383       &   \cellcolor[HTML]{FFEFEF}0.8663    \\
&   & \cellcolor[HTML]{F3FFF2}$E_{\xi}^{mean}\uparrow$ &  \cellcolor[HTML]{F3FFF2}0.8792     &    \cellcolor[HTML]{F3FFF2}0.8051  &  \cellcolor[HTML]{F3FFF2}0.8921    &  \cellcolor[HTML]{F3FFF2}0.8344   &     \cellcolor[HTML]{F3FFF2}0.7901  &    \cellcolor[HTML]{F3FFF2}0.8877  &        \cellcolor[HTML]{F3FFF2}0.6758   &      \cellcolor[HTML]{F3FFF2}0.7937      &      \cellcolor[HTML]{F3FFF2}0.6250     &         \cellcolor[HTML]{F3FFF2}0.9220      &    \cellcolor[HTML]{F3FFF2}0.9434                     \\\cdashline{2-14}[5pt/3pt]
& \multirow{4}{*}{Average}    & \cellcolor[HTML]{DCF7FF}$MAE\downarrow$ &    \cellcolor[HTML]{DCF7FF}0.0352 &   \cellcolor[HTML]{DCF7FF}0.0419 &  \cellcolor[HTML]{DCF7FF}0.0355    &   \cellcolor[HTML]{DCF7FF}0.0763  &   \cellcolor[HTML]{DCF7FF}0.0366    &    \cellcolor[HTML]{DCF7FF}0.0393  &    \cellcolor[HTML]{DCF7FF}0.0815       &     \cellcolor[HTML]{DCF7FF}0.0440       &     \cellcolor[HTML]{DCF7FF}0.1098      &    \cellcolor[HTML]{DCF7FF}0.0239           &       \cellcolor[HTML]{DCF7FF}0.0197   \\
&      & \cellcolor[HTML]{FFFFE0}$S\uparrow$ & \cellcolor[HTML]{FFFFE0}0.8605      &  \cellcolor[HTML]{FFFFE0}0.8232    &  \cellcolor[HTML]{FFFFE0}0.8579    &   \cellcolor[HTML]{FFFFE0}0.8166  &  \cellcolor[HTML]{FFFFE0}0.8442     &  \cellcolor[HTML]{FFFFE0}0.8520    & \cellcolor[HTML]{FFFFE0}0.7182    &  \cellcolor[HTML]{FFFFE0}0.8207   &  \cellcolor[HTML]{FFFFE0}0.6943    &  \cellcolor[HTML]{FFFFE0}0.8998  &   \cellcolor[HTML]{FFFFE0}0.9114   \\
&      &  \cellcolor[HTML]{FFEFEF}$F_{\beta}^{mean}\uparrow$ & \cellcolor[HTML]{FFEFEF}0.7876      &  \cellcolor[HTML]{FFEFEF}0.7282    &  \cellcolor[HTML]{FFEFEF}0.7827    &  \cellcolor[HTML]{FFEFEF}0.7274   &  \cellcolor[HTML]{FFEFEF}0.7604     &  \cellcolor[HTML]{FFEFEF}0.7708    &  \cellcolor[HTML]{FFEFEF}0.5541         &   \cellcolor[HTML]{FFEFEF}0.7213         &    \cellcolor[HTML]{FFEFEF}0.5404       &  \cellcolor[HTML]{FFEFEF}0.8457       &   \cellcolor[HTML]{FFEFEF}0.8611    \\
&   & \cellcolor[HTML]{F3FFF2}$E_{\xi}^{mean}\uparrow$ &  \cellcolor[HTML]{F3FFF2}0.8804     &    \cellcolor[HTML]{F3FFF2}0.8294  &  \cellcolor[HTML]{F3FFF2}0.8771    &  \cellcolor[HTML]{F3FFF2}0.8342   &     \cellcolor[HTML]{F3FFF2}0.8638  &    \cellcolor[HTML]{F3FFF2}0.8747  &        \cellcolor[HTML]{F3FFF2}0.7029   &      \cellcolor[HTML]{F3FFF2}0.8268      &      \cellcolor[HTML]{F3FFF2}0.6859     &         \cellcolor[HTML]{F3FFF2}0.9261      &    \cellcolor[HTML]{F3FFF2}0.9379                     \\\hline     
\end{tabular}
}
\end{table*}

To verify that training models with images containing various weather noises yields better robustness than training with clean images or images with a single noise type serves as a prerequisite for this study on leveraging noise characteristics to enhance model segmentation performance under complex weather conditions. Thus, seven models (each composed of the conventional ResNet-50 and decoder) are trained using different training subsets. The experimental results are presented in Table~\ref{tab:quantitative_comparison_subsets}, where the second-to-last row “Multi-weather” indicates that the seven methods are trained using the complete WXSOD training set~(100\% training data), and the last row “Multi-weather (w.NIFM)” denotes that the seven methods further adopt the specific encoder enhanced by NIFM. Due to space limitations, we will subsequently focus analysis primarily on the average results.

\begin{figure*}[!t]
\centering
\includegraphics[width=1.0\linewidth]{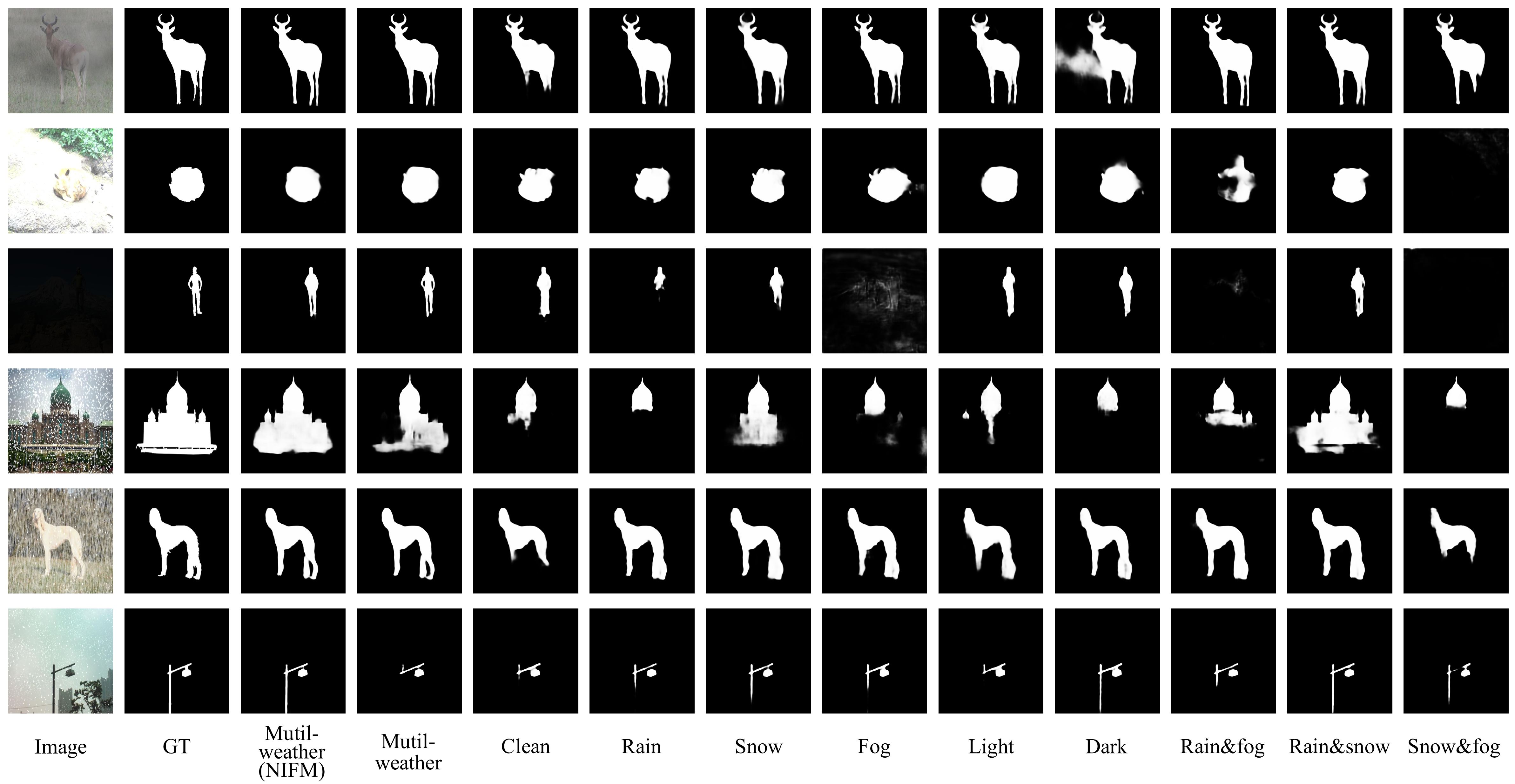}
\caption{Visual Comparison on synthesized test set across different training subsets.}
\label{sys_subset_visual}
\end{figure*}  

\begin{figure*}[!t]
\centering
\includegraphics[width=1.0\linewidth]{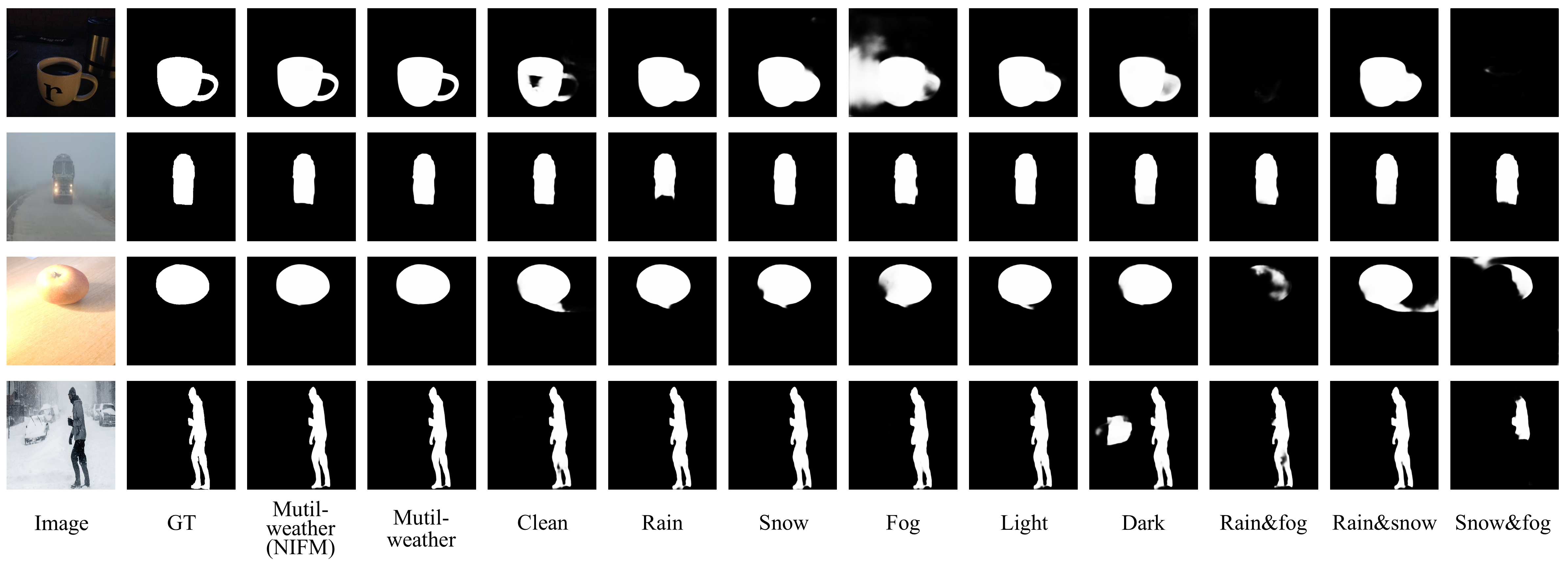}
\caption{Visual Comparison on real test set across different training subsets.}
\label{real_subset_visual}
\end{figure*}  

From Table~\ref{tab:quantitative_comparison_subsets}, it can be observed that models trained with the larger-scale and noise-diverse mixed training set (\textit{i.e.}, "Multi-weather") can significantly improve segmentation performance, with all evaluation metrics enhanced. On the other hand, models trained on subsets with a single noise type~(\textit{e.g}, "Dark", "Fog" and "Snow\&Fog") generally perform worse in all metrics compared to those trained on the "Clean" subset. This is mainly because the test set contains complex noise types, and using a training set with a single noise type instead enlarges the domain gap between training and test data, undermining the model’s generalization ability. Secondly, in terms of various evaluation metrics of both the Synthesized test set and the Real test set, the model represented by “Multi-weather (w.NIFM)” performs significantly better than the model corresponding to “Multi-weather”. Taking the average MAE metric as an example, on the Synthesized test set, the average MAE of “Multi-weather (w.NIFM)” is 0.0315, which is lower than 0.0339 of “Multi-weather”; on the Real test set, the average MAE of the former is 0.0197, also significantly lower than 0.0239 of the latter. For other metrics such as $S$ and $F_\beta^{adp}$, “Multi-weather~(w.NIFM)”-based models also achieves the optimal results of most key metrics. This demonstrates that enhancing the specific encoder via NIFM can effectively integrate multi-type weather noise information and significantly improve the model’s segmentation performance for salient objects in complex weather scenarios, further verifying the effectiveness of the method proposed in this paper in utilizing noise differences.

We also provide visual comparisons of the BAFSNet model under different training subset settings, as shown in Fig.~\ref{sys_subset_visual} and Fig.~\ref{real_subset_visual}. In contrast to training subsets with a single noise type, BAFSNet, when trained under the “Multi-weather” setting with richer training data, exhibits more precise segmentation results. Moreover, upon integrating NIFM into the conventional encoder utilized in BAFSNet based on the “Multi-weather” setup, the segmentation accuracy is further improved.

\section{Conclusion}\label{Conclusions}
This paper focuses on the SOD under complex weather-corrupted scenarios, which represents a practical yet understudied task. Most existing SOD methods are designed under the clean-image assumption, and their performance tends to degrade under adverse weather conditions. 
To address this limitation, we propose a SOD framework consisting of a weather-aware encoder and a replaceable decoder, which can be seamlessly integrated with other SOD models. The core of this work lies in a novel Noise Indicator Fusion Module~(NIFM). NIFM takes both semantic features and a one-hot noise indicator as inputs to adaptively learn noise-type-dependent feature weights, thereby performing secondary modulation on the semantic features extracted by the encoder. To ensure comprehensive feature refinement, we embed one NIFM between every two consecutive stages of the encoder, enabling progressive modulation of multi-scale semantic features. Extensive experiments are conducted on the WXSOD dataset, covering multiple training data scales (100\%, 50\%, and 30\% of the full training set) and seven distinct decoder configurations. Results demonstrate that, compared with conventional encoders, the NIFM-enhanced encoder achieves improved detection accuracy under complex weather conditions. Future work will explore extending the noise indicator to more continuous or implicit representations, and validate the framework's effectiveness on a wider ranger of unstructured degradations.


 

%

\bibliographystyle{IEEEtran}
\bibliography{ref.bib}

\vfill

\end{document}